\documentclass[preprint,12pt]{elsarticle}

\usepackage{amssymb}
\usepackage{amsmath}
\usepackage{amsmath}
\usepackage{xcolor}   % 颜色支持
\usepackage{cancel}   % 删除线支持
\usepackage{ifthen}   % 条件判断

\newboolean{showmarkup}
\setboolean{showmarkup}{false} % true=标记版本 / false=干净版本

\newcommand{\add}[1]{% 新增内容
    \ifthenelse{\boolean{showmarkup}}
        {\textcolor{red}{#1}}   % 标记版本：红色显示
        {#1}                    % 干净版本：直接显示
}

\journal{Mechanism and Machine Theory}

\begin{document}

\begin{frontmatter}



\title{Kinetostatics and Particle-Swarm Optimization of Vehicle-Mounted Underactuated Metamorphic Loading Manipulators} %% Article title


\author[1,2]{Nan Mao}
\author[1]{Junpeng Chen\corref{cor1}} 
\ead{chenjp2020@mail.sustech.edu.cn}
        \author[3]{Guanglu Jia\corref{cor1}}
        \ead{jiagl@simtech.a-star.edu.sg}     	
	\author[2]{Emmanouil Spyrakos-Papastavridis}
	\author[1,2]{Jian S. Dai}
       \cortext[cor1]{Corresponding author}

        %% Author name
\affiliation[1]{organization={Institute for Robotics Research}, 
addressline={Southern University of Science and Technology}, 
city={Shenzhen}, 
postcode={518055}, 
country={China}}
  
\affiliation[2]{organization={Centre for Robotics Research}, 
addressline={King’s College London}, 
city={London}, 
postcode={WC2R 2LS}, 
country={United Kingdom}}

\affiliation[3]{organization={Singapore Institute of Manufacturing Technology, Agency for Science, Technology and Research (A*STAR)}, 
city={Singapore}, 
postcode={138635}, 
country={Singapore}}

%% Abstract
\begin{abstract}
Fixed degree-of-freedom (DoF) loading mechanisms often suffer from excessive actuators, complex control, and limited adaptability to dynamic tasks. This study proposes an innovative mechanism of underactuated metamorphic loading manipulators (UMLM), integrating a metamorphic arm with a passively adaptive gripper. The metamorphic arm exploits geometric constraints, enabling the topology reconfiguration and flexible motion trajectories without additional actuators. The adaptive gripper, driven entirely by the arm, conforms to‌ diverse objects through passive compliance. A structural model is developed, and a kinetostatics analysis is conducted to investigate isomorphic grasping configurations. To optimize performance, Particle-Swarm Optimization (PSO) is utilized to refine the gripper’s dimensional parameters, ensuring robust adaptability across various applications. Simulation results validate the UMLM's easily implemented control strategy, operational versatility, and effectiveness in grasping diverse objects in dynamic environments. This work underscores the practical potential of underactuated metamorphic mechanisms in applications requiring efficient and adaptable loading solutions. Beyond the specific design, this generalized modeling and optimization framework extends to a broader class of manipulators, offering a scalable approach to the development of robotic systems that require efficiency, flexibility, and robust performance.

\end{abstract}

%% Keywords
\begin{keyword}
kinetostatics analysis, metamorphic manipulator, underactuated mechanism, Particle-Swarm Optimization, optimization methodology


\end{keyword}

\end{frontmatter}



\section{Introduction}
\label{sec1}
%% Labels are used to cross-reference an item using \ref command.

\add{With the increasing integration of robotics into human society, the demand for advanced robotic manipulator systems has grown significantly due to the need for increased flexibility and efficiency \cite{c75}. Emulating the fundamental functions of the human hands, robotic manipulator systems are extensively utilized across various domains such as aerospace and manufacturing \cite{c1}.} \add{The development of robotic hands has seen remarkable progress in mechanical structure design and control strategies. Structural innovations aim to enhance flexibility and adaptability, with many scholars contributing significantly to the field \cite{c2}, by developing devices such as the Utah/MIT hand \cite{c3}, TUAT/Karlsruhe hand \cite{c4}, and BCL-13 \cite{c5}. Control innovations focus on improving precision, responsiveness, and autonomy, including model-based, analytical, optimization-based, and learning-based methods \cite{c86}. In recent years, learning-based approaches have attracted widespread attention from researchers due to their efficiency in handling complex manipulation tasks. Notable examples include Corona et al. and Lundell et al. \cite{c87, c88}.}

\add{Traditional manipulator systems face challenges due to their numerous actuation components and intricate control systems, making it difficult to balance the trade-offs between achieving task flexibility \cite{c6, c76} and versatility \cite{c7, c77}.} \add{Driven by these limitations,  flexible manipulators are characterized by high dexterity, light weight, and low stiffness, attracting scholars from various countries to conduct extensive research \cite{c8, c93, c94, c95}.} Typical flexible manipulators are exemplified by the flexible humanoid hand developed by Ohio State University in the United States \cite{c9}, the Octopus Gripper produced by Festo in Germany \cite{c10}, and the three-finger pneumatic soft manipulator created by Ritsumeikan University in Japan \cite{c11}. Pisa/IIT SoftHand is one of the most popular soft robotic hands with 19 joints using only one actuator to activate its adaptive synergy \cite{c58}. In China, typical soft manipulators consist of the pneumatic soft manipulator from Shanghai Jiao Tong University \cite{c12}, the four-finger soft manipulator of Beihang University \cite{c13}, and the two-finger soft manipulator from National Cheng Kung University in Taiwan, China \cite{c14}.
	
\add{However, the challenge of excessive weight and cost is pronounced due to the need for individual actuators to power each joint's degree of freedom, along with sophisticated control and sensing systems \cite{c15}. These limitations have motivated the pursuit of manipulators that can perform multi-tiered grasping actions \cite{c25} and adaptively handle items of various shapes and sizes \cite{c26} with simplified control\cite{c16}, leading to the development of underactuated manipulator systems. Underactuated systems possess fewer actuators than degrees of freedom \cite{c23}, utilizing a hybrid actuation strategy that combines active motors with passive elements like springs or coupling mechanisms to achieve grasping capabilities \cite{c24}. This has effectively solved high integration costs of conventional systems and spurred research into examples such as the SDM \cite{c17}, PASA-GB hand \cite{c18}, and SARAH \cite{c19,c20,c21,c22}.}
	
Metamorphic mechanisms have the characteristics of variable mobility and topology \cite{c29, c30}. They can self-assemble and reconfigure \cite{c31, c60}, changing their configuration during operation suddenly from plural to singular \cite{c32}, adapting the loading manipulators to the task requirements of different work stages \cite{c61}, and maximizing its functionality \cite{c33}. In unstructured environments, this adaptability is crucial for efficient grasping and manipulation, prompting the innovation and marking a significant advancement in dexterity and functionality \cite{c27,c28}. Originating from biological terminology, metamorphosis describes the process of transformation and structural adaptation \cite{c34, c62, c63, c89}. As applied to robotics, Dai introduced the concept of the metamorphic mechanism in 1998 \cite{c35, c36}. Such mechanisms are reconfigurable \cite{c37}, enabling variable mobility and topology for the robots during operation \cite{c90}, which greatly enhances their functionality and adaptability \cite{c38}. Furthermore, several methodologies for synthesis and configuration design of metamorphic mechanisms have been developed \cite{c39, c40, c41, c42}. \add{The synthesis of metamorphic mechanisms is one of the core challenges in the theoretical study due to the complexity of developing methods that address variable constraints and dynamic transformations across different functional states \cite{c64}. Three main types of configuration synthesis methods have been developed for metamorphic mechanisms \cite{c65}. The first approach relies on mathematical representations, such as adjacency matrices and configuration graphs, to describe and manipulate topological changes \cite{c66, c78}. Representative works include matrix-based synthesis methods and topology modeling approaches \cite{c67, c79}. Tian et al. proposed a novel synthesis of reconfigurable generalized parallel mechanisms (GPMs) using kinematotropic linkages and configurable platforms, with kinematic constraints analyzed through Lie group theory \cite{c80}. The second type constructs new metamorphic mechanisms by replacing conventional joints with metamorphic kinematic joints such as rT joint \cite{c68}, rR joint \cite{c69} and vA joint \cite{c70}. The third type is task-oriented, focusing on the variation of constraint characteristics based on functional demands. This method formulates metamorphic equations using working-phase and source matrices, enabling systematic synthesis based on constraint variation and integration of sub-configurations, which is particularly promising for practical applications due to its strong adaptability to task-specific requirements. Several researchers have contributed to the development of task-oriented synthesis methods for metamorphic mechanisms. Wang et al. \cite{c71} introduced the concepts of working-phase and source-metamorphic mechanisms. Zhang et al. \cite{c72} analyzed  variations in constraints arising from multiple configurations and proposed a synthesis approach based on the coupling of adjacent configurations using metamorphic joints. Chang et al. \cite{c73} applied constraint screw theory to describe the relationship between source and sub-configurations, formulating a variable-constraint synthesis method. Li et al. \cite{c74} further developed synthesis methods using equivalent resistance and constraint structure matrices to characterize metamorphic joints.} Building on these concepts, a range of metamorphic mechanisms have been introduced, encompassing those with adjustable topologies \cite{c43} and multi-modal capabilities \cite{c44}. \add{Notable examples include the origami-inspired integrated 8R kinematotropic metamorphic mechanism \cite{c45} and planar five-bar metamorphic linkage with five phases resulting from locking of motors proposed by Tian et al. \cite{c81}.} This design incorporates mathematical and topological insights \cite{c46} to achieve greater dexterity \cite{c47}. Origaker is a novel multi-mimicry quadruped robot able to transform between different working modes, as the reptile-, arthropod-, and mammal-like modes without disassembly and reassembly, bridging the gap between biological metamorphosis and robotic metamorphosis \cite{c59}. \add{Additionally, recent developments in variable stiffness manipulators have demonstrated significant potential for enhancing robotic adaptability and safety in unstructured environments \cite{c82}. By dynamically adjusting joint stiffness, these systems enable compliant interactions with unpredictable surroundings, which is crucial for human–robot collaboration and tasks involving physical contact \cite{c83}. Cheng et al. \cite{c84} applied the reversible jamming of granular media and vacuum pressure to adjust the stiffness of the system. Shiva et al. \cite{c85} proposed a continuum silicon-based variable stiffness manipulator using an antagonistic actuation scheme. D. J. Braun et al. \cite {c91} proposed a framework for optimizing torque and impedance profiles in anthropomorphic robots with variable-impedance actuators, addressing real-world actuation constraints to maximize task performance.}

Combining the advantages of underactuation and metamorphosis, this study creates a novel mechanism of underactuated metamorphic loading manipulators (UMLM) that features a streamlined design and enhanced versatility with coupling sub-chains \cite{c92}.
	
Particle-Swarm Optimization (PSO) is one of the most well-regarded swarm-based algorithms \cite{c48} introduced by Kennedy and Eberhart \cite{c49}, and has been applied in a host of areas such as mechanism optimization \cite{c50}, parameter selection, and structural topology optimisation \cite{c51, c52}. In recent years, robotics has witnessed a significant adoption of intelligent PSO algorithms for parameter optimization \cite{c53}. Gong et al. \cite{c54} developed a hybrid optimization algorithm combining bee swarm and PSO to optimize manipulators parameters. Wang et al. \cite{c55} introduced a universal index with an enhanced PSO for precise pose selection. Fang and Dang \cite{c56} presented a QPSO method for kinematic calibration across serial and parallel robots, enhancing calibration accuracy. These contributions epitomise the evolving landscape of parameter identification in robotics \cite{c57}.
	
This research presents a study on kinematics, statics modeling analysis, and parameter optimization using the PSO Algorithm, for the purpose of determining the best parameters of the UMLM. A validated simulation model supports the effectiveness of the proposed system. The principal contributions are as follows:
	
(1) A novel metamorphic topology mechanism is proposed for the UMLM system, enabling adaptive underactuated manipulation. Driven by a single motor, this mechanism supports multifunctional operations—grasping, releasing, lifting, and descending, allowing the gripper to conform to objects of varying shapes and sizes.

\add{(2) A comprehensive kinetostatic modeling of the underactuated gripper mechanisms is established to analyze the structural behavior and finger–object contact forces. This kinetostatic-based contact force analysis enables the derivation of optimal finger joint lengths based on the geometric attributes of target objects, thereby facilitating scalable customization of the manipulation systems for real-world applications.}

(3) A tailored Particle-Swarm Optimization (PSO) algorithm is developed to enhance the gripper’s structural efficiency and parameter sensitivity. This method optimizes the structural parameters such that, under limited actuation, the contact forces are uniformly distributed among the finger joints, significantly improving grasping performance under dynamic loads and diverse object geometries.

(4) Beyond the proposed prototype, the modeling and optimization methodology can be extended to a broader class of metamorphic and underactuated manipulators, supporting future development of adaptable robotic systems in unstructured environments.

The flowchart in Fig.~\ref{figure1} displays a comprehensive overview of the methodology for the UMLM design and optimization as proposed in this paper. We developed an innovative mechanism of UMLM, detailing its metamorphic mechanism and underactuation to enhance adaptability. A mathematical model was constructed to capture the manipulator's kinematics and statics, forming the basis for subsequent analyses. kinetostatics-based analysis across various motion modes was conducted to optimize task adaptation and transition smoothness. Contact force distribution on the gripper was analyzed using static models and virtual work principles, which are crucial for manipulator grasping stability and performance. An objective function for gripping performance was formulated, focusing on uniform contact force distribution, with design variables identified for optimization. The PSO algorithm was utilized to refine the gripper's structural parameters. Finally, simulations validated the UMLM's performance in terms of grasping objects of diverse sizes and shapes. The remainder of this paper is organized as follows.
    \add{
	\begin{figure}[htbp]
		\centering 	\includegraphics[width=1\linewidth]{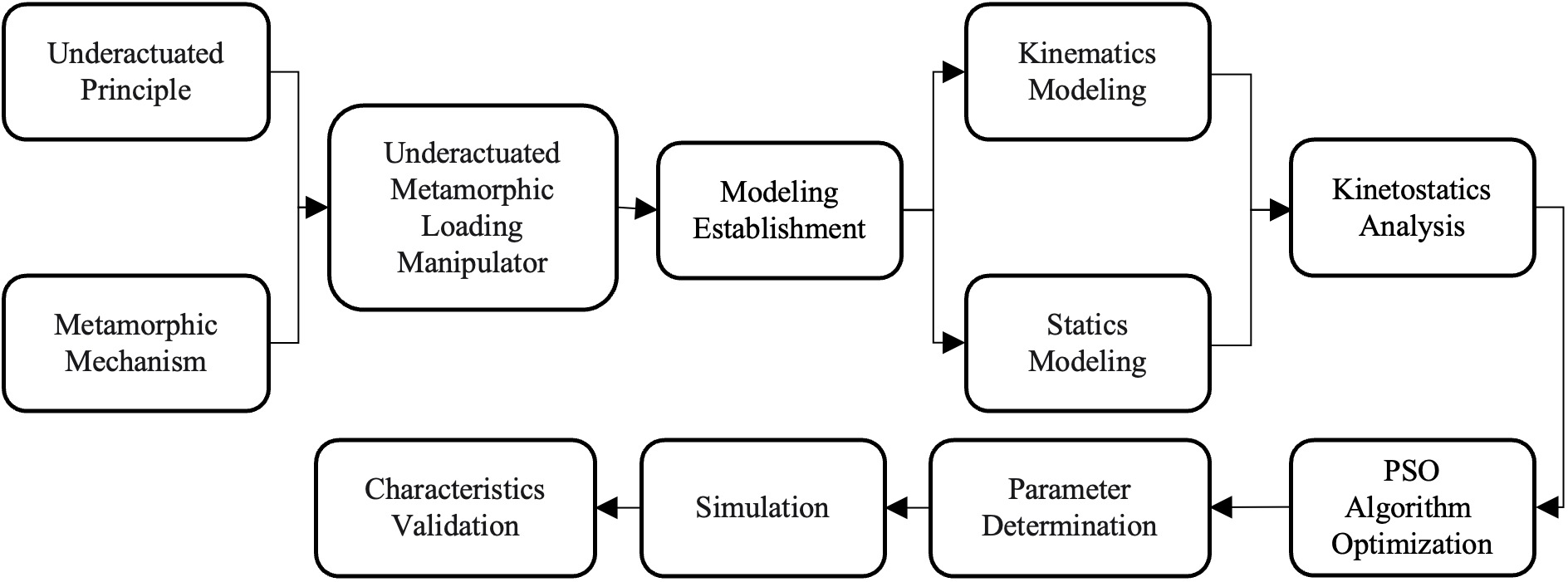}
		\add{\caption{Optimization methodology diagram of the UMLM design.}}
		\label{figure1}
	\end{figure}
}
 
 Following the introduction, In Section 2, the overall structural model of the UMLM is outlined and the kinematics analysis of the UMLM is conducted. \add{Section 3 demonstrates the kinetostatics-based analysis of coordinated motion for the UMLM on a movable vehicle. Then, statics modeling of the UMLM's grasping is established and kinetostatics-based contact force analysis is conducted in Section 4.}The finger's structural parameters have been meticulously adjusted to generate ample gripping force, ensuring reliable control over objects even with a restricted actuation capacity. Furthermore, to promote a more consistent distribution of contact forces among the fingers and the grasped object, a Particle-Swarm Optimization (PSO) algorithm is utilized to refine the gripper’s parameter settings in Section 5. Section 6 exams the various isomorphic configurations of the manipulator's fingers during object grasping through a simulation model of the manipulator grasping tests using objects of varying sizes and shapes to validate the proposed strategies, followed by an analysis of the simulation results. Lastly, the conclusion is presented in Section 7.	

\add{\section{Kinematics Modeling and Analysis via the Closed-loop Method of the UMLM}}

\subsection{The Metamorphosis Property of the Mechanism}

\add{In this study, we propose a novel design for the UMLM as shown in Fig.~\ref{figure2}. The 3D structural mechanism diagram of the UMLM is illustrated in Fig.~\ref{figure2}(a), which provides a detailed representation of the manipulator’s constituent elements and their interrelationships. The UMLM is designed to execute a variety of fundamental tasks, which mainly consists of an underactuated robotic arm part and an underactuated gripper part. The underactuated robotic arm is facilitated by two distinct topological configuration transformations, including grasping-releasing mode and lifting-descending mode, with only one actuating element. The underactuated gripper facilitates seamless interaction with objects of varying shapes and sizes, allowing the UMLM system to adaptively execute grasping and releasing motions.}
\add{
\begin{figure}[htbp]
\centering
\includegraphics[width=14cm]{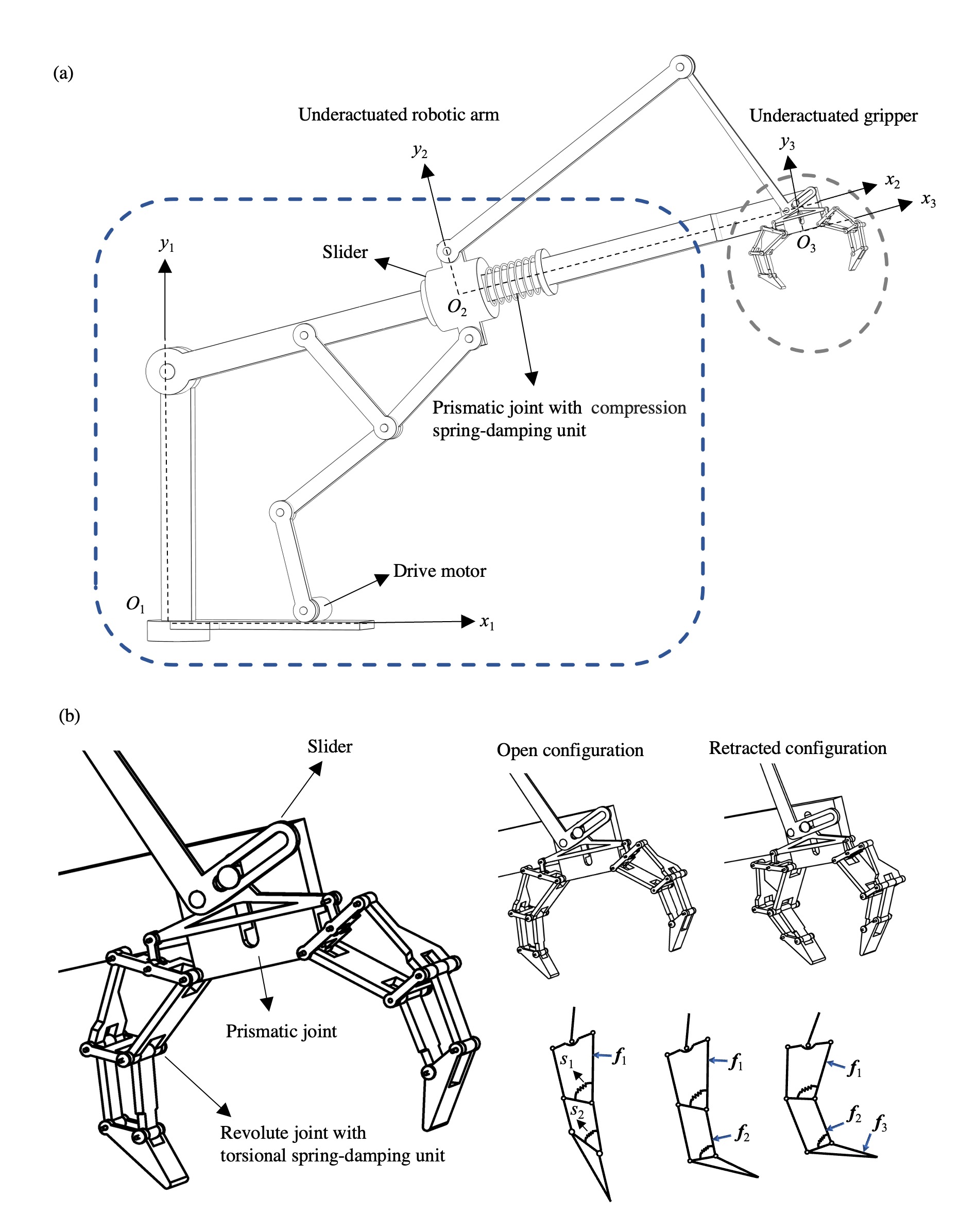}
  \add{\caption{3D Diagram of the UMLM metamorphic mechanism. (a) Geometric definitions of the 3D UMLM model. (b) Geometric definitions and metamorphic process of the gripper model.}
  \label{figure2}}
\end{figure}
}
The detailed structural mechanism diagram of the metamorphic gripper is demonstrated in Fig.~\ref{figure2}(b). Each metamorphic finger mechanism consists of two torsional springs, two four-bar linkages and one three-bar linkage, forming three distinct finger segments as illustrated in Fig.~\ref{figure2}(b). Each four-bar mechanism incorporates a torsional spring at the revolute joint designed to prevent internal collapse and facilitate sequential closure during grasping. Upon contact with the object, the first finger joint engages spring $s_1$, restricting its motion and preventing relative movement between the joints, which causes the first four-bar linkage to behave as a rigid body. Subsequently, the second finger is driven by the input torque to rotate. When the object contacts the second finger joint, spring $s_2$ is engaged, limiting relative motion and causing the second four-bar linkage to become a rigid body. The contact forces between the inner sides of the three finger joints and the grasped object are \(\boldsymbol{f}_1\), \(\boldsymbol{f}_2\), and \(\boldsymbol{f}_3\). This process continues until all three finger joints have made contact with the object, or the mechanical limits of the mechanism are reached as illustrated in Fig.~\ref{figure2}(b).

\begin{figure}[htbp]
		\centering 	\includegraphics[width=13cm]{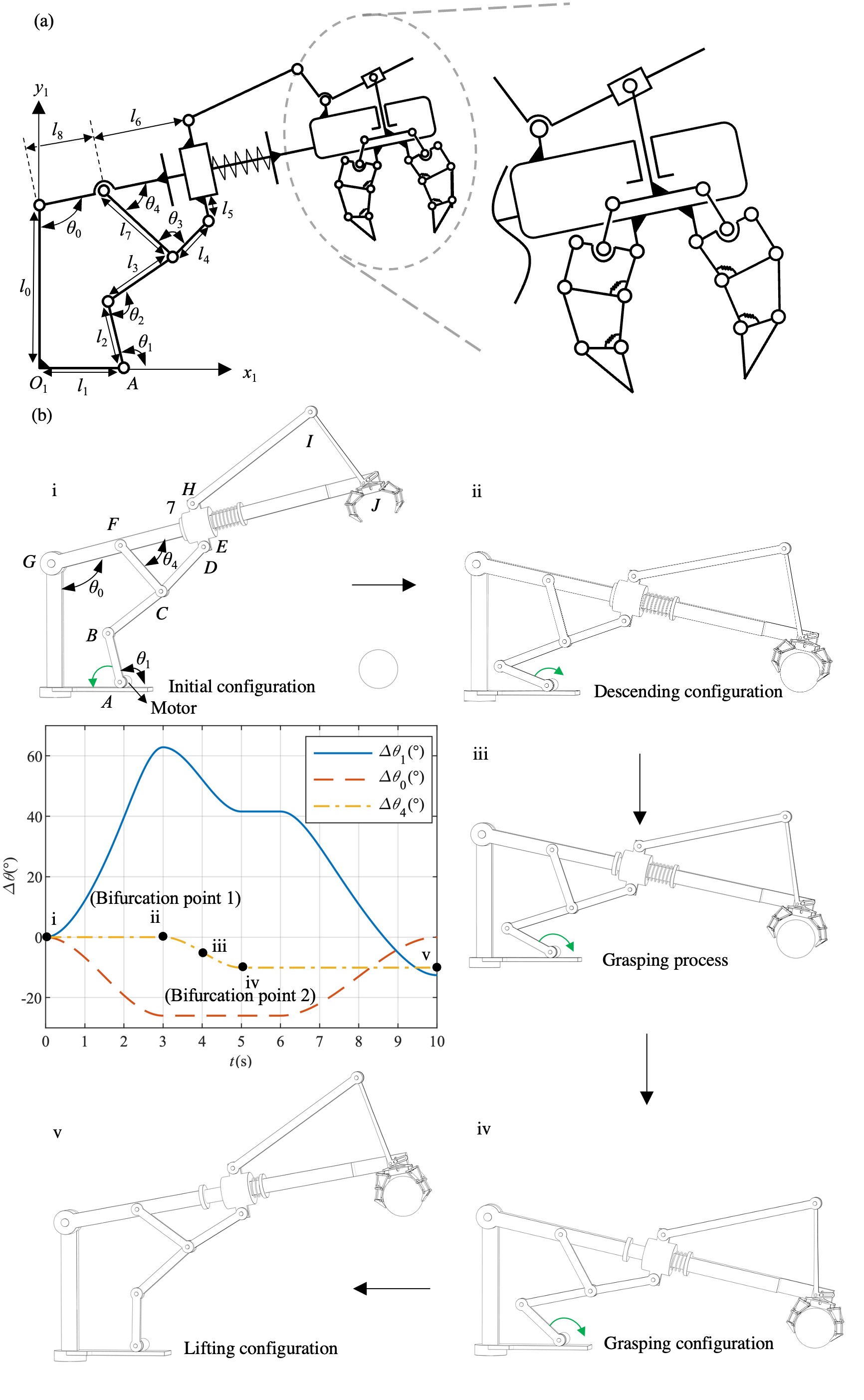}	\caption{Metamorphic process and kinematic variations of the angles in the loading process. (a) Schematic diagram of the UMLM metamorphic mechanism for loading. (b) Metamorphic process of different configurations and kinematic angle variations curves of \(\theta_1\), \(\theta_0\), \(\theta_4\) in the loading process.}
		\label{figure3}
	\end{figure}

The schematic diagram of the UMLM is demonstrated in Fig.~\ref{figure3}(a). The metamorphic process of the proposed UMLM robotic arm is demonstrated in Fig.~\ref{figure3}(b), which illustrates different phases of a typical grasping action performed by the UMLM and the corresponding trajectories of \( \theta_1 \), \( \theta_0 \) and \( \theta_4 \). The simulation parameter settings are demonstrated in Table~\ref{tab:simulation_params}. The topological change of the robotic arm is achieved through the geometric limitation of the slider. When the slider is restrained, the actuator can make the entire robotic arm perform lifting and descending movements, as shown in Figs.~\ref{figure3}(b)i and v. When the slider is released, the lateral movement of the slider can control the opening and closing of the adaptive gripper, as shown in Figs.~\ref{figure3}(b)ii, iii and iv. Meanwhile, the adaptive gripper, as an underactuated mechanism, can adapt to grasp different objects through the geometric limitation of the torsional spring-damping unit. 

The grasping process is as follows: from the initial configuration of Fig.~\ref{figure3}(b)i, the slider 7 is held at the leftmost position under the spring force. When the motor drives joint \( A \) to rotate counterclockwise, \( \theta_1 \) increases and \( \theta_0 \) decreases, thereby initiating the downward motion of the manipulator to approach the object. When the gripper descends to the target position as shown in Fig.~\ref{figure3}(a)ii, it is in the pre-grasping state. When the motor drives joint \( A \) to rotate clockwise with \( \theta_1 \) decreasing, the slider 7 is released and moves rightward along the rod while the transmitted motor force exceeds the spring force (with \( \theta_4 \) decreasing while \( \theta_0 \) remains constant), which controls the adaptive gripper to close around the object, as demonstrated in Figs.~\ref{figure3}(b)iii and iv. Once the grasping is completed, slider 7 reaches the rightmost end and becomes locked in place. Even though the spring force acts to push the slider leftward, the continued motor rotation provides sufficient force to maintain the slider’s position against the spring force. Simultaneously, the gripper maintains its grasp on the object with \( \theta_4 \) remaining constant. When the torque of the motor continues to increase while the motor continues rotating clockwise (with \( \theta_1 \) decreasing), the entire manipulator is lifted as in Fig.~\ref{figure3}(b)v. 

As shown, the lifting-descending and grasping-releasing processes are achieved with only one actuator at joint \( A \). The corresponding kinematic behaviors of the configuration transformations can be seen in Fig.~\ref{figure3}(b), demonstrating the angle variations during different configuration transformations. The variation in $\Delta \theta_1$, characterized by an initial increase followed by a decrease, corresponds to the driving motor first rotating counterclockwise and then clockwise. A change in $\Delta \theta_0$ represents the manipulator's lifting or lowering motion. Furthermore, the variation in $\Delta \theta_4$ indicates slider translation, which dictates the grasping or releasing action of the gripper.

\begin{table}[t]
\centering

\caption{Dynamic parameters of simulation experiments.}\label{tab:simulation_params}

\begin{tabular}{l c}
\hline
Parameter & Numerical values\\
\hline
Slider spring stiffness \( k \) (N/mm) & \( 4.2 \) \\
Torsional spring \( s_1 \) stiffness \( k_1 \) (Nmm/rad) & \( 346.5\) \\
Torsional spring \( s_2 \) stiffness \( k_2 \)  (Nmm/rad) & \( 794.1\) \\
DC motor (W) & 120 \\
Force exponent \( \epsilon \) (mm) & 1.50 \\
Static friction coefficient \( \mu_s \) & 0.2 \\
Dynamic friction coefficient \( \mu_d \) & 0.1 \\
\( l_0 \) (mm) & 397 \\
\( l_1 \) (mm) & 181 \\
\( l_2 \) (mm) & 130 \\
\( l_3 \) (mm) & 180 \\
\( l_4 \) (mm) & 160 \\
\( l_5 \) (mm) & 58 \\
Mass of the UMLM system (kg) & 44.5\\
Mass of the sphere (kg) & 0.52 \\
Diameter of the sphere (mm) & 100 \\
Mass of the cylinder (kg) & 4.94 \\
Diameter of the cylinder (mm) & 100 \\
Mass of the pentagonal prism (kg) & 0.95 \\
Side length of pentagonal prism (mm) & 64 \\
Mass of the irregular object (kg) & 0.65 \\
Compression torque of the spring (Nm) & 12.6 \\
Torque required to lift the robotic arm (Nm) & 30.5 \\
\hline
\end{tabular}

\end{table}

\begin{figure}[htbp]
  \centering   \includegraphics[width=13.16cm]{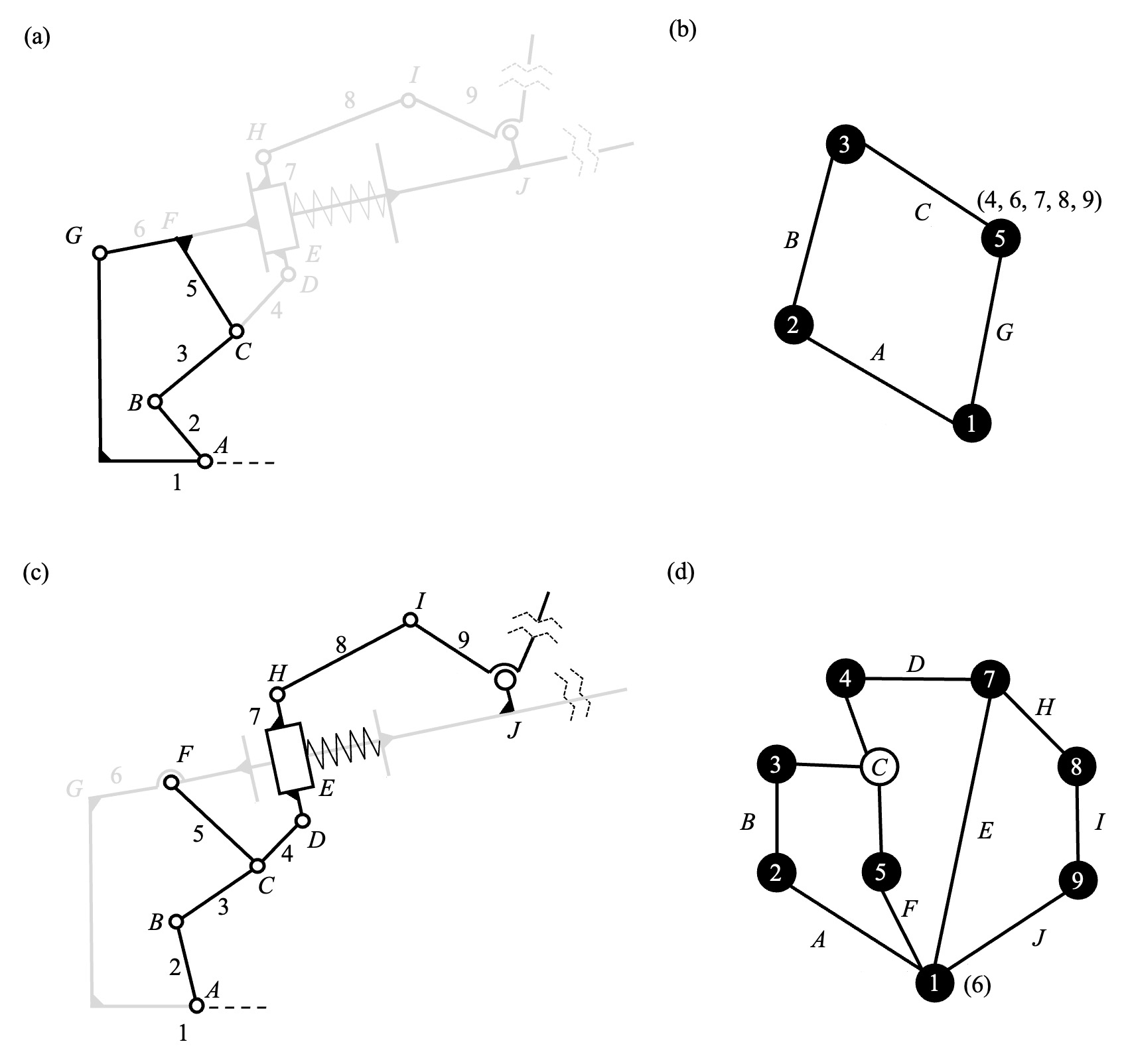}
  \caption{Diagram of metamorphosis property and topological phase change of the UMLM. (a) and (b) are the schematic diagram and corresponding topological diagram of the topology 1 (lifting or descending configuration). (c) and (d) are the schematic diagram and corresponding topological diagram of the topology 2 (grasping or releasing configuration). Black circles 1–9 denote the links, white circle \(C\) represents the composite hinge, letters \(A\)–\(J\) denote the kinematic pairs.}
  \label{figure4}
\end{figure}

The schematic and topological diagram of different configurations are illustrated in Fig.~\ref{figure4}. When the slider 7 moves to either of the left or right limit position, the four-bar linkage \(EHIJ\) becomes a rigid body, rendering the gripper immobile. At the same time, the mechanism degenerates into a four-bar linkage \(ABCG\), representing topology 1 of the mechanism with one degree of freedom. This topology 1 is illustrated in Figs.~\ref{figure4}(a) and (b), which ensure the vertical motion capabilities of the UMLM, enabling precise and stable lifting and descending actions. When performing the grasping and releasing operations, the mechanism maintains a consistent topological configuration, corresponding to topology 2 as illustrated in Figs.~\ref{figure4}(c) and (d). When the motor begins to rotate clockwise, the transmitted motor force exceeds the spring force and drives the slider 7 rightward along the rod, this makes the gripper to grasp an object. The robotic arm consists of a six-bar linkage \(ABCDEF\) combined with a four-bar linkage \(EHIJ\), resulting in a total of one degree of freedom. 

This chapter provides an introduction to the conceptual framework and mechanical innovations that underpin the design and functionality of the UMLM, highlighting the system's transformative properties. Through this exploration, the metamorphic capability of the UMLM is revealed, emphasizing its metamorphosis characteristics in performing a range of actions through topological configuration transformations. The design not only addresses common limitations of fixed-DoF manipulators, but also provides a scalable foundation for future applications in unstructured and dynamic environments.
	
\add{\subsection{Kinematics Modeling via the Closed-loop Method of the UMLM}}

The structure of the robotic arm is illustrated in Fig.~\ref{figure3}(a). It consists of rods and slides. The system's torque input is applied at joint \(A\).

\add{\(l_i\) represents the rod length, \(i=0,1,2,...,8\), \(\theta_0\) represents the angle between the arm rod \(l_0\) and vertical rod \(l_1\), \(\theta_1\) represents the angle of the drive, \(\theta_2\) represents the angle between rod \(l_2\) and \(l_3\), \(\theta_4\) represents the angle between rod \(l_6\) and \(l_7\), \(\omega_i\) and \(\beta_i\) represent the corresponding angular velocity and angular acceleration of \(\theta_i\).} It is assumed that throughout the operation of the robotic arm, the arm maintains a constant speed.
 
\add{A Cartesian coordinate system $\{O_{1}-x_1y_1z_1\}$ is established, with its origin at point \(O_{1}\) located at the base of the underactuated metamorphic loading manipulator. The \(x_1\)-axis is aligned with the direction of \(l_{1}\), the \(y_1\)-axis is aligned with the direction of \(l_{0}\). The closed-loop equation is formulated as follows:}
	\begin{equation}
    \add{
		\boldsymbol{l}_{2} + \boldsymbol{l}_{3} + \boldsymbol{l}_{7} = \boldsymbol{l}_{1} + \boldsymbol{l}_{0} + \boldsymbol{l}_{8}.
        }
	\end{equation}
	
The \add{closed-loop} equation is projected onto the x and y directions, yielding the following position equations:
	
	\begin{equation}
		\begin{aligned}
			x: \quad l_2 \cos \theta_1 - l_3 \cos (\theta_1 + \theta_2) - l_7 \sin (\theta_0 - \theta_4) &= -l_1 + l_8 \sin \theta_0,\\
			y: \quad l_2 \sin \theta_1 - l_3 \sin (\theta_1 + \theta_2) + l_7 \cos (\theta_0 - \theta_4) &= l_0 - l_8 \cos \theta_0.
		\end{aligned}
  \label{eq2}
	\end{equation}
	
Based on the above analysis, the relationships among \(\theta_0\), \(\theta_1\) and \(\theta_4\) under the existing dimensional parameters are illustrated. The above expressions form a nonlinear system of equations. 

By fixing \(\theta_0\) or \(\theta_4\), the motion of the robotic arm can be categorized into two phases: the lifting process and the grasping process.
	
\add{During the lifting process, \(\theta_4\) is fixed. The angles \(\theta_0\) and \(\theta_2\) can be determined by \(\theta_1\) through geometric relationships in Eq. \ref{eq2}.}By differentiating the displacement, the velocity expressions are obtained as follows:
	
	\begin{equation}
		\left[\begin{array}{cc}
			l_7 \cos \left(\theta_0-\theta_4\right)+l_8 \cos \theta_0 & -l_3 \sin \left(\theta_1+\theta_2\right) \\
			l_7 \sin \left(\theta_0-\theta_4\right)+l_8 \sin \theta_0 & l_3 \cos \left(\theta_1+\theta_2\right)
		\end{array}\right]\left[\begin{array}{l}
			\omega_0 \\
			\omega_2
		\end{array}\right]=\left[\begin{array}{c}
			-l_2 \sin \theta_1+l_3 \sin \left(\theta_1+\theta_2\right) \\
			l_2 \cos \theta_1-l_3 \cos \left(\theta_1+\theta_2\right)
		\end{array}\right] \omega_1.
	\end{equation}
	
\add{During the grasping process, \(\theta_0\) is fixed. The angles \(\theta_4\) and \(\theta_2\) can be determined by \(\theta_1\) through the geometric relationships in Eq. \ref{eq2}.} By differentiating the displacement, the velocity expressions are obtained as follows: 

	\begin{equation}
		\left[\begin{array}{cc}
			-l_3 \sin \left(\theta_1+\theta_2\right) & -l_7 \cos \left(\theta_0-\theta_4\right) \\
			l_3 \cos \left(\theta_1+\theta_2\right) & -l_7 \sin \left(\theta_0-\theta_4\right)
		\end{array}\right]\left[\begin{array}{l}
			\omega_2 \\
			\omega_4
		\end{array}\right]=\left[\begin{array}{c}
			-l_2 \sin \theta_1+l_3 \sin \left(\theta_1+\theta_2\right) \\
			l_2 \cos \theta_1-l_3 \cos \left(\theta_1+\theta_2\right)
		\end{array}\right] \omega_1.
	\end{equation}
	
Substituting the obtained angles $\theta_0$, $\theta_1$, $\theta_2$, $\theta_4$ into equation (4) yields the angular velocity of the UMLM system.

Since the motion maintains a constant speed, $\frac{\mathrm{d} \omega_1}{\mathrm{~d} t}=0$. Differentiating this relationship yield the acceleration relationships as follows:
	
	\begin{equation}
		\begin{aligned}
			& {\left[\begin{array}{cc}
					-l_3 \sin \left(\theta_1+\theta_2\right) & -l_7 \cos \left(\theta_0-\theta_4\right) \\
					l_3 \cos \left(\theta_1+\theta_2\right) & -l_7 \sin \left(\theta_0-\theta_4\right)
				\end{array}\right]\left[\begin{array}{l}
					\beta_2 \\
					\beta_4
				\end{array}\right]} \\
			& =\left[\begin{array}{ll}
				-l_2 \omega_1 \cos \theta_1+l_3\left(\omega_1+\omega_2\right) \sin \left(\theta_1+\theta_2\right) \\
				-l_2 \omega_1 \sin \theta_1-l_3\left(\omega_1+\omega_2\right) \cos \left(\theta_1+\theta_2\right)
			\end{array}\right] \omega_1 \\
			& -\left[\begin{array}{ll}
				-l_3\left(\omega_1+\omega_2\right) \cos \left(\theta_1+\theta_2\right) & -l_7 \omega_4 \sin \left(\theta_0-\theta_4\right) \\
				-l_3\left(\omega_1+\omega_2\right) \sin \left(\theta_1+\theta_2\right) & l_7 \omega_4 \cos \left(\theta_0-\theta_4\right)
			\end{array}\right]\left[\begin{array}{l}
				\omega_2 \\
				\omega_4
			\end{array}\right].
		\end{aligned}
	\end{equation}
	
Substituting the obtained angles $\theta_0$, $\theta_1$, $\theta_2$, $\theta_4$ and angular velocities $\omega_1$, $\omega_2$, $\omega_4$ into equation (5) yields the angular acceleration of the UMLM system. 

From the geometric relationships, the displacement of the slider is related to \(\theta_4\) as follows:

	\begin{equation}
		l_6=l_7 \cos \theta_4+l_4 \sin \left(\cos ^{-1}\left(\frac{l_7 \sin \theta_4-l_5}{l_4}\right)\right).
	\end{equation}
	
Differentiating yields the following relationship between the velocity of the slider and  \(\theta_4\) and \(\omega_4\) :
	
	\begin{equation}
		\frac{\mathrm{d} l_6}{\mathrm{~d} t}=-l_7 \omega_4 \sin \theta_4+\frac{\left(l_7 \omega_4 \cos \theta_4\right)\left(l_5-l_7 \sin \theta_4\right)}{l_4}\left(\frac{1}{\sqrt{1-\left(\frac{\left(l_7 \sin \theta_4-l_5\right)}{l_4}\right)^2}}\right).
	\end{equation}
	
Differentiating again yields the following relationship between the acceleration of the slider and \(\theta_4\), \(\omega_4\), and \(\beta_4\):
\begin{equation}
\begin{aligned}
\frac{\mathrm{d}^2 l_6}{\mathrm{~d} t^2} &= -l_7\left(\beta_4 \sin \theta_4 + \omega_4^2 \cos \theta_4\right)
+ \frac{c}{\sqrt{b}} \cdot \frac{l_7\left(\beta_4 \cos \theta_4 - \omega_4^2 \sin \theta_4\right)}{l_4} \\
&\quad - \frac{a}{\sqrt{b}} \cdot \frac{l_7^2 \omega_4^2}{l_4} \left(1 + \frac{c^2}{b l_4^2}\right),
\end{aligned}
\end{equation}
\noindent{\text{where} \quad}\\
\begin{equation}
a = \cos^2 \theta_4, \\
b = 1 - \frac{c^2}{l_4^2}, \\
c = l_5 - l_7 \sin \theta_4.
\nonumber
\end{equation}

\add{At this stage, the kinematic relationships for \(\theta_1\), \(\theta_0\), \(\theta_4\), as well as for the robotic arm have been established.}
\subsection{The Coupling Relationship between Gripper Input and the UMLM}

\add{In this segment of our scientific investigation, we assume that the gripper section of the UMLM receives an input force that is perpendicular to the upper arm via virtue of the intermediary of a slider mechanism, allowing us to decouple the complex dynamics of the gripper from the broader analysis of the robotic arm, thereby facilitating a more focused and detailed examination of each component's behavior. The decoupling of the gripper analysis enables us to isolate the forces and movements acting upon the gripper, which is crucial for understanding its performance and optimizing its design.}

\add{Our subsequent analysis will delve into the intricate relationship between the slider and the vertical input force exerted on the gripper, which directly influences the gripper's ability to securely grasp and manipulate objects with precision. We will scrutinize how the slider's position and movement are correlated with the vertical input, and how these factors contribute to the overall performance of the gripper.}

\add{The $\{O_{2}-x_2y_2z_2\}$ coordinate system is established, with its origin at point \(O_2\) representing the slider of the UMLM. The \(x_2\)-axis is aligned with the direction of \(l_{9}\), the \(y_2\)-axis is aligned with the direction perpendicular to \(l_{9}\).} The relevant symbol conventions are illustrated in Fig.~\ref{figure6}. The \add{closed-loop} equation is formulated as follows:

    	\begin{figure}[htbp]
		\centering 
		\includegraphics[width=10.3cm]{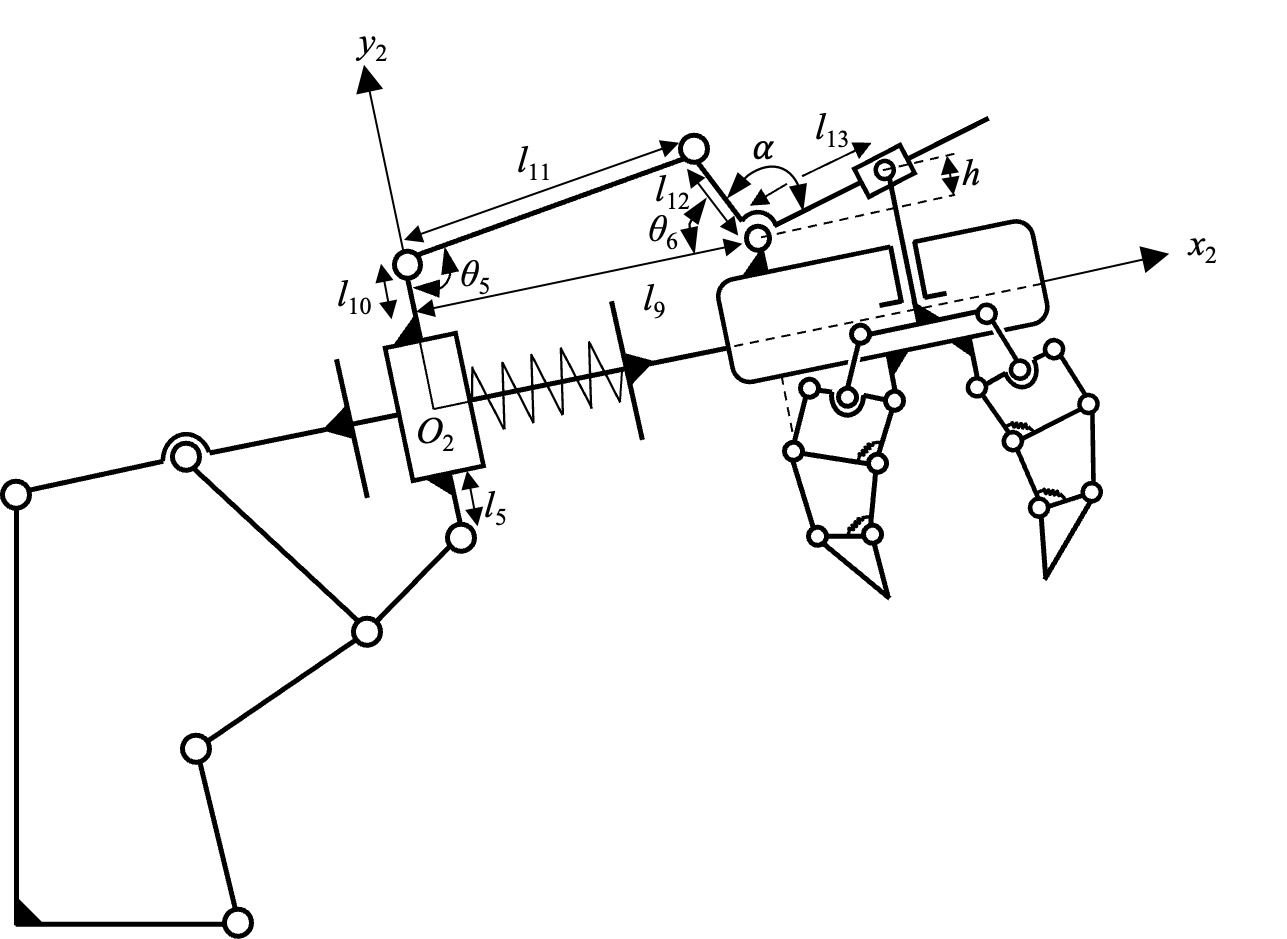}
		\caption{Schematic diagram of the slider metamorphic mechanism.}
		\label{figure6}
	\end{figure}

	\begin{equation}
\add{
\boldsymbol{l}_{10}+\boldsymbol{l}_{11}=\boldsymbol{l}_{9}+\boldsymbol{l}_{12}.\\
}
	\end{equation}
	
By projecting this \add{closed-loop} equation onto the  \(x\) and \(y\) directions, the positional relationships are obtained as follows:
	
	\begin{equation}
    \add{	
		\begin{array}{ll}
			x: & 0+l_{11} \cos \left(\theta_5-\frac{\pi}{2}\right)=l_9+l_{12} \cos \left(\pi-\theta_6\right), \\
			y: & l_{10}+l_{11} \sin \left(\theta_5-\frac{\pi}{2}\right)=0+l_{12} \sin \left(\pi-\theta_6\right).
		\end{array}
        }
	\end{equation}
	
These equations form a nonlinear system. Given \(l_9\), angles \(\theta_5\) and \(\theta_6\) can be solved through geometric relationships. 
Differentiating yields the velocity relationships:

	\begin{equation}
		\left[\begin{array}{cc}
			l_{11} \cos \theta_5 & -l_{12} \sin \theta_6 \\
			l_{11} \sin \theta_5 & -l_{12} \cos \theta_6
		\end{array}\right]\left[\begin{array}{l}
			\omega_5 \\
			\omega_6
		\end{array}\right]=\left[\begin{array}{l}
			1 \\
			0
		\end{array}\right] \frac{\mathrm{d} l_9}{\mathrm{~d} t}.
	\end{equation}

Differentiating again provides the acceleration relationships:

	\begin{equation}
		\left[\begin{array}{cc}
			l_{11} \cos \theta_5 & -l_{12} \sin \theta_6 \\
			l_{11} \sin \theta_5 & -l_{12} \cos \theta_6
		\end{array}\right]\left[\begin{array}{l}
			\beta_5 \\
			\beta_6
		\end{array}\right]=\left[\begin{array}{l}
			1 \\
			0
		\end{array}\right] \frac{\mathrm{d}^2 l_9}{\mathrm{~d} t^2}-\left[\begin{array}{cc}
			-l_{11} \omega_5 \sin \theta_5 & -l_{12} \omega_6 \cos \theta_6 \\
			l_{11} \omega_5 \cos \theta_5 & l_{12} \omega_6 \sin \theta_6
		\end{array}\right]\left[\begin{array}{l}
			\omega_5 \\
			\omega_6
		\end{array}\right].
	\end{equation}

From the geometric relationships, $\theta_6$ is related to $h$ as follows:
\begin{equation}
\add{
        h=l_{13} \sin \left({\pi}-\theta_6-\alpha\right).
        }
	\end{equation}

Differentiating yields the velocity relationships:

	\begin{equation}
    \add{
	\frac{\mathrm{d} h}{\mathrm{~d} t}=-l_{13} \omega_6\cos\left({\pi}-\theta_6-\alpha\right).
    }
	\end{equation}

Differentiating again provides the acceleration relationships:

	\begin{equation}
    \add{
	\frac{\mathrm{d}^2 h}{\mathrm{~d} t^2}=-l_{13} \omega_6^2 \sin \left(\theta_6+\alpha\right)+l_{13}\beta_6\cos\left(\theta_6+\alpha\right).
        }
	\end{equation}
\add{At this stage, the kinematic relationships of the UMLM as well as the gripper input have been established.}
\subsection{Kinematics of the UMLM with Cross-section to the Gripper during Stable Grasping}
	
\begin{figure}[htbp]
		\centering 
		\includegraphics[width=9.22cm]{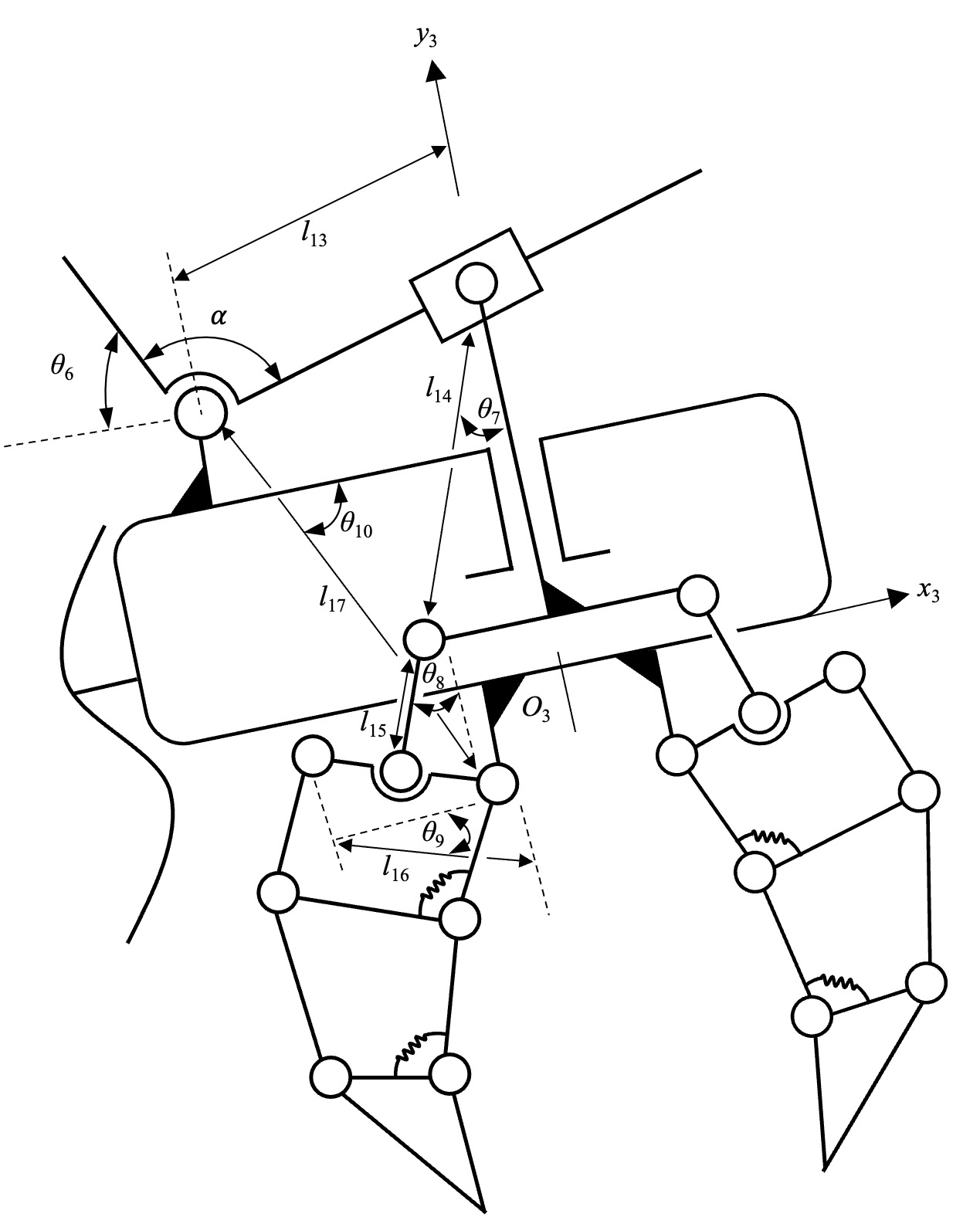}
		\add{\caption{Schematic diagram of cross-section to the gripper.}
		\label{figure7}}
	\end{figure}

The vertical input mechanism is illustrated in Fig.~\ref{figure7}. \add{The \(\{O_{3}-x_3y_3z_3\}\) coordinate system is established, with its origin at point \(O_{3}\) denoting the lower midpoint of the approximated prismatic pair of the gripper. The \(x_3\)-axis is aligned with the direction perpendicular to \(l_{9}\), the \(y_3\)-axis is aligned with the direction perpendicular to \(l_{9}\).} Other symbol conventions are as shown in Fig.~\ref{figure7}. The \add{closed-loop} equation is formulated as follows:
	\begin{equation}
\add{		\boldsymbol{l}_{13}+\boldsymbol{l}_{14}+\boldsymbol{l}_{15}+\frac{\boldsymbol{l}_{16}}{2}=\boldsymbol{l}_{17}.
}
	\end{equation}
    
By projecting this \add{closed-loop} equation onto the \(x\) and \(y\) directions, the positional relationships are obtained as follows:
	\begin{equation}
    \add{	
		\begin{array}{ll}
			x: &l_{13}\cos\left({\pi}-\theta_6-\alpha\right)-l_{14} \sin \theta_7-l_{15} \sin \theta_8+\frac{l_{16}}{2} \cos \theta_9=l_{17} \cos \theta_{10}, \\
			y: & l_{13} \sin\left({\pi}-\theta_6-\alpha\right)-l_{14} \cos \theta_7-l_{15} \cos \theta_8-\frac{l_{16}}{2} \sin \theta_9=-l_{17} \sin \theta_{10}.
		\end{array}
        }
	\end{equation}
    
Differentiating yields the velocity relationships:
	\begin{equation}
        \add{	
		\left[\begin{array}{lllll}
			l_{13} \sin\left({\pi}-\theta_6-\alpha\right) & -l_{14} \cos \theta_7 & -l_{15} \cos \theta_8& -\frac{l_{16}}{2} \sin \theta_9& l_{17} \sin \theta_{10} \\
			-l_{13} \cos\left({\pi}-\theta_6-\alpha\right) & l_{14} \sin \theta_7 & l_{15} \sin \theta_8& -\frac{l_{16}}{2} \cos \theta_9& l_{17} \cos \theta_{10}
		\end{array}\right]\left[\begin{array}{l}
			\omega_6 \\
            \omega_7 \\
            \omega_8 \\
			\omega_9\\
            \omega_{10}
		\end{array}\right]=0.
        }
	\end{equation}	
	
Differentiating again provides the acceleration relationships:
\add{
\begin{equation}
\begin{aligned}
\left[\begin{array}{lllll}
    l_{13} \sin\!\left({\pi}-\theta_6-\alpha\right) & -l_{14} \cos \theta_7 & -l_{15} \cos \theta_8 & -\frac{l_{16}}{2} \sin \theta_9 & l_{17} \sin \theta_{10} \\
    -l_{13} \cos\!\left({\pi}-\theta_6-\alpha\right) & l_{14} \sin \theta_7 & l_{15} \sin \theta_8 & -\frac{l_{16}}{2} \cos \theta_9 & l_{17} \cos \theta_{10}
\end{array}\right]
\!\!
\begin{bmatrix}
    \beta_6 \\ \beta_7 \\ \beta_8 \\ \beta_9 \\ \beta_{10}
\end{bmatrix}
&\\
=\left[\begin{array}{lllll}
    l_{13} \cos\!\left({\pi}-\theta_6-\alpha\right) & -l_{14} \sin \theta_7 & -l_{15} \sin \theta_8 & \frac{l_{16}}{2} \cos \theta_9 & -l_{17} \cos \theta_{10} \\
    l_{13} \sin\!\left({\pi}-\theta_6-\alpha\right) & - l_{14} \cos \theta_7 & - l_{15} \cos \theta_8 & -\frac{l_{16}}{2} \sin \theta_9 & l_{17} \sin \theta_{10}
\end{array}\right]
\begin{bmatrix}
    \omega^2_6 \\ \omega^2_7 \\ \omega^2_8 \\ \omega^2_9 \\ \omega^2_{10}
\end{bmatrix}.
\end{aligned}
\end{equation}
}
    	
Through this detailed analysis, we aim to uncover the underlying principles governing the interaction between the slider and the gripper, and how these principles can be harnessed to enhance the manipulator's functionality. 

\add{\subsection{Kinematics Analysis of the Underactuated Gripper}}

\begin{figure}[htbp]
		\centering 
\includegraphics[width=11.5cm]{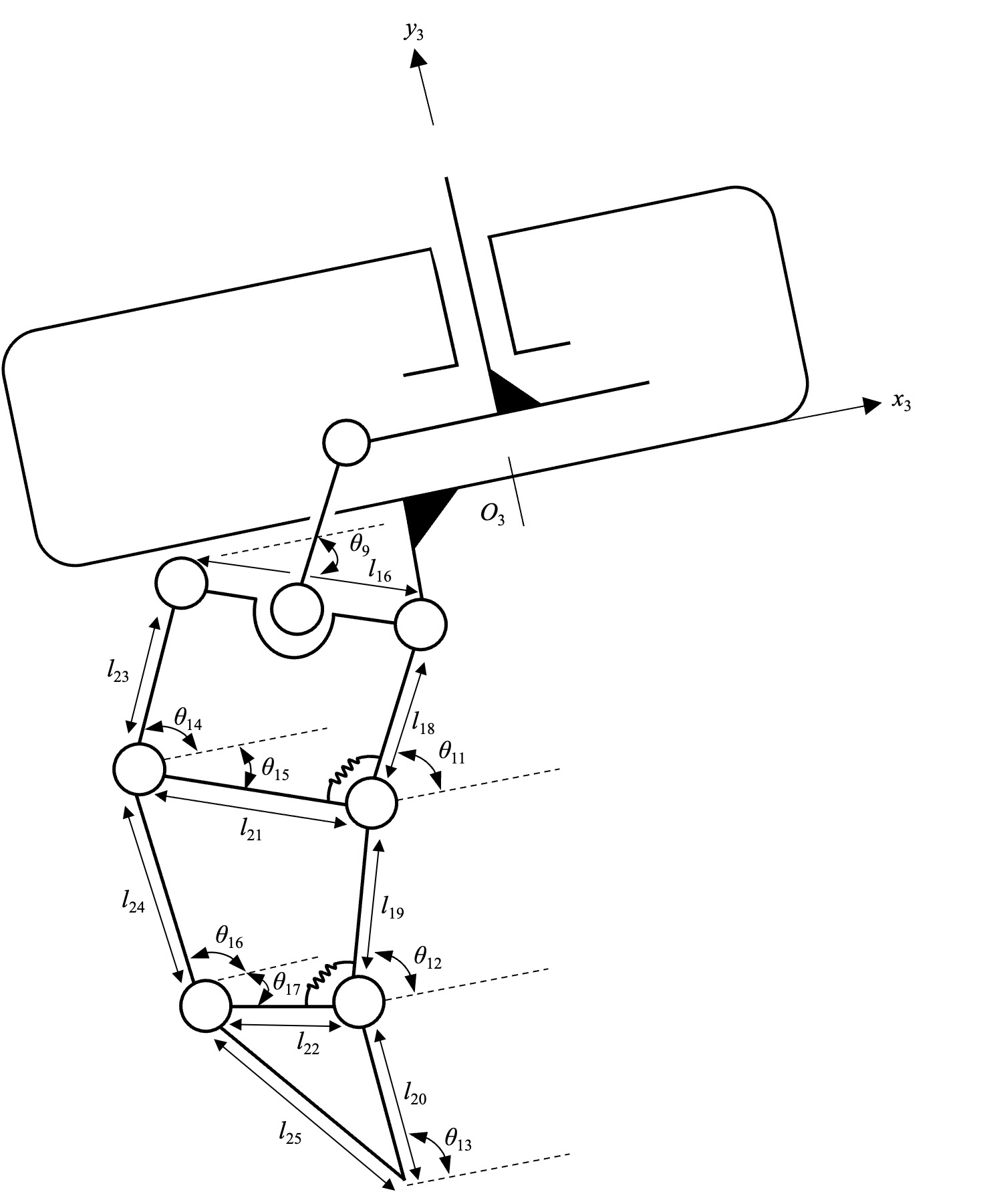}
\add{\caption{Schematic diagram of the underactuated gripper.}
		\label{add1}}
	\end{figure}

\add{The underactuated gripper mechanism is illustrated in Fig.~\ref{add1}. The closed-loop equation of the first finger joint mechanism is formulated as follows:
	\begin{equation}		\boldsymbol{l}_{16}+\boldsymbol{l}_{23}=\boldsymbol{l}_{18}+\boldsymbol{l}_{21}.
    \end{equation}
}
	
\add{By projecting this onto the \(x_3\) and \(y_3\) directions, we have:
	\begin{equation}
		\begin{array}{ll}
			x: &l_{16}\cos\theta_{9}+l_{23}\cos\theta_{14}-l_{21}\cos\theta_{15}-l_{18}\cos\theta_{11}=0, \\
			y: & l_{16}\sin\theta_{9}+l_{23}\sin\theta_{14}-l_{21}\sin\theta_{15}-l_{18}\sin\theta_{11}=0.
		\end{array}
	\end{equation}
    }

\add{By differentiating we have:
\begin{equation}
\begin{aligned}
x: & \quad -l_{16}\sin \theta_{9}\omega_{9} - l_{23}\sin \theta_{14}\omega_{14} + l_{21}\sin \theta_{15}\omega_{15} + l_{18}\sin \theta_{11}\omega_{11} = 0, \\
y: & \quad l_{16}\cos \theta_{9}\omega_{9} + l_{23}\cos \theta_{14}\omega_{14} - l_{21}\cos \theta_{15}\omega_{15} - l_{18}\cos \theta_{11}\omega_{11} = 0.
\end{aligned}
\label{eq5}
\end{equation}
}

\add{The closed-loop equation of the second finger joint mechanism is formulated as follows:
	\begin{equation}		\boldsymbol{l}_{21}+\boldsymbol{l}_{24}=\boldsymbol{l}_{19}+\boldsymbol{l}_{22}.
    \end{equation}
}

\add{By projecting this onto the \(x_3\) and \(y_3\) directions, we have:
	\begin{equation}
		\begin{array}{ll}
			x: &l_{21}\cos\theta_{15}+l_{24}\cos\theta_{16}-l_{22}\cos\theta_{17}-l_{19}\cos\theta_{12}=0, \\
			y: & l_{21}\sin\theta_{15}+l_{24}\sin\theta_{16}-l_{22}\sin\theta_{17}-l_{19}\sin\theta_{12}=0.
		\end{array}
	\end{equation}
    }

\add{By differentiating we have:
	\begin{equation}
    		\begin{array}{ll}
			x:
&-l_{21}\sin \theta_{15}\omega_{15}-l_{24}\sin \theta_{16}\omega_{16}+l_{22}\sin \theta_{17}\omega_{17}+l_{19}\sin \theta_{12}\omega_{12}= 0,\\
y: & 
l_{21}\cos \theta_{15}\omega_{15}+l_{24}\cos \theta_{16}\omega_{16}-l_{22}\cos \theta_{17}\omega_{17}-l_{19}\cos \theta_{12}\omega_{12}= 0.
		\end{array}
    \end{equation}
}

\add{The third finger joint is constructed as a fixed 3-bar linkage to reduce the number of unknowns. Hence, the geometric relationship is given as:}
	\add{
    \begin{equation}
    \cos(\theta_{13}+\theta_{17})=\frac{{l_{20}}^2+{l_{22}}^2-{l_{25}}^2}{2{l_{20}}{l_{22}}}
     \end{equation}
}

\add{Up to now, we have completed the analysis of the kinematic relationships within the entire UMLM system. Once the position and size of the grasped object are obtained, we can calculate the joint  \(A\)ngles based on the contact point coordinates.}

\begin{figure}[!htbp]
		\centering 
		\includegraphics[width=12.89cm]{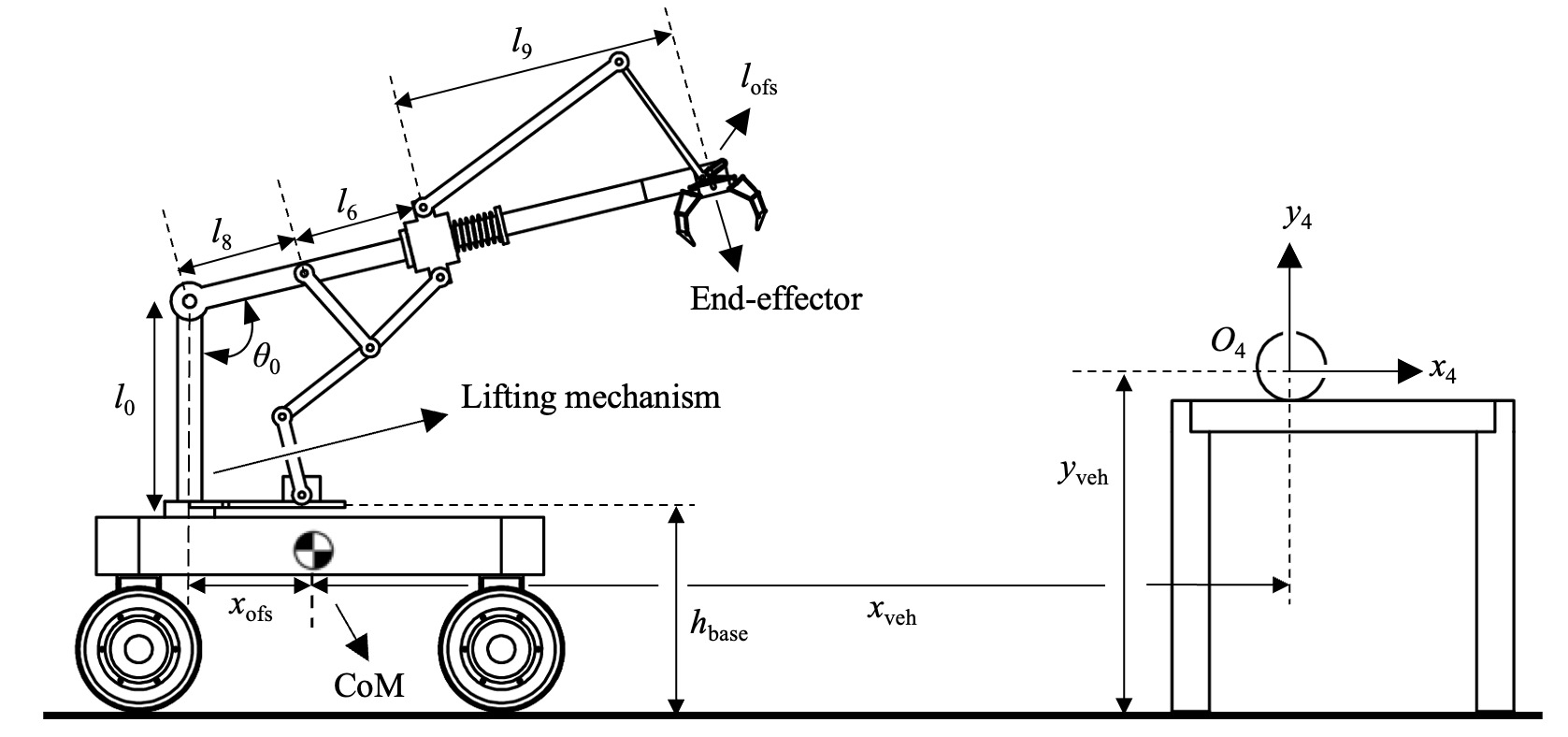}
		\add{\caption{The underactuated mechanical arm is installed on a mobile vehicle.}
		\label{figure8}}
	\end{figure}

\add{\section{Kinetostatics-based Analysis of Coordinated Motion for the UMLM on a Movable Vehicle}}
The kinetostatics of the UMLM was modelled in the preceding sub-sections. The UMLM is a single degree of freedom mechanism, resulting in a unique and coupled end-effector (EE) trajectory, which inherently limits its operational workspace. To expand its applicability to a wider range of scenarios, the lifting function is integrated with the UMLM system through mounting atop mobile platforms, as illustrated in Fig. \ref{figure8}. \add{The center of mass (CoM) of the movable vehicle is illustrated in Fig.~\ref{figure8}. The $\{O_{4}-x_4y_4z_4\}$ coordinate frame is established with its origin at the center $O_{4}$ of the grasped object. The \(x_4\)-axis is aligned with the direction of \(OA\), the \(y_4\)-axis is aligned with the direction perpendicular to \(OA\).}

 The end-effector gripper exhibits self-adaptive capabilities, enabling grasping of simple objects based solely on positional alignment without requiring precise orientation adjustments. The end-effector trajectory depends exclusively on the angle $\theta_0$ of the primary joint of the robotic arm. As derived in Fig.~\ref{figure3} and Eq. \ref{eq2}, which establishes the relationship between the actively driven angle $\theta_1$ and $\theta_0$, we directly express the end-effector pose using $\theta_0$. The grasping position coordinates are defined as:  

 \begin{equation}
     \begin{aligned}
EE_x &= -x_{\mathrm{veh}} - x_{\mathrm{ofs}} + \left(l_8 + l_6 + l_9 + l_{\mathrm{ofs}}\right) \cos\left(\theta_0 - \frac{\pi}{2}\right), \\
EE_y &=-y_{\mathrm{veh}} +h_{\mathrm{base}}  + l_0 + \left(l_8 + l_6 + l_9 + l_{\mathrm{ofs}}\right) \sin\left(\theta_0 - \frac{\pi}{2}\right),
     \end{aligned}\label{eq3}
 \end{equation}
here $x_{\mathrm{veh}}$ is the horizontal distance between the movable vehicle and the target object, $y_{\mathrm{veh}}$ is the height of the target vehicle, $h_{\mathrm{base}}$ is the height of the UMLM base.

Here, $x_{{ofs}}$ represents the manipulator's mounting offset. During coordinated grasping, $\theta_0$ can first be determined from $EE_y$ using:

\begin{equation}
\theta_0 = \sin^{-1}\left(\frac{EE_y +y_{\mathrm{veh}}- h_{\mathrm{base}} - l_0}{l_8 + l_6 + l_9 + l_{\mathrm{ofs}}}\right) + \frac{\pi}{2}.
\end{equation}

Once $\theta_0$ is obtained, the height $\Delta h$ of the change in the lifting mechanism can be determined according to Eq. \ref{eq3}. Meanwhile, the required horizontal displacement of the movable vehicle is calculated using $EE_x$:

\begin{equation}
x_{\mathrm{veh}} = -EE_x - x_{\mathrm{ofs}} + \left(l_8 + l_6 + l_9 + l_{\mathrm{ofs}}\right) \cos\left(\theta_0 - \frac{\pi}{2}\right).
\label{eq22}
\end{equation}

\begin{figure}[!htbp]
		\centering
		\includegraphics[width=14cm]{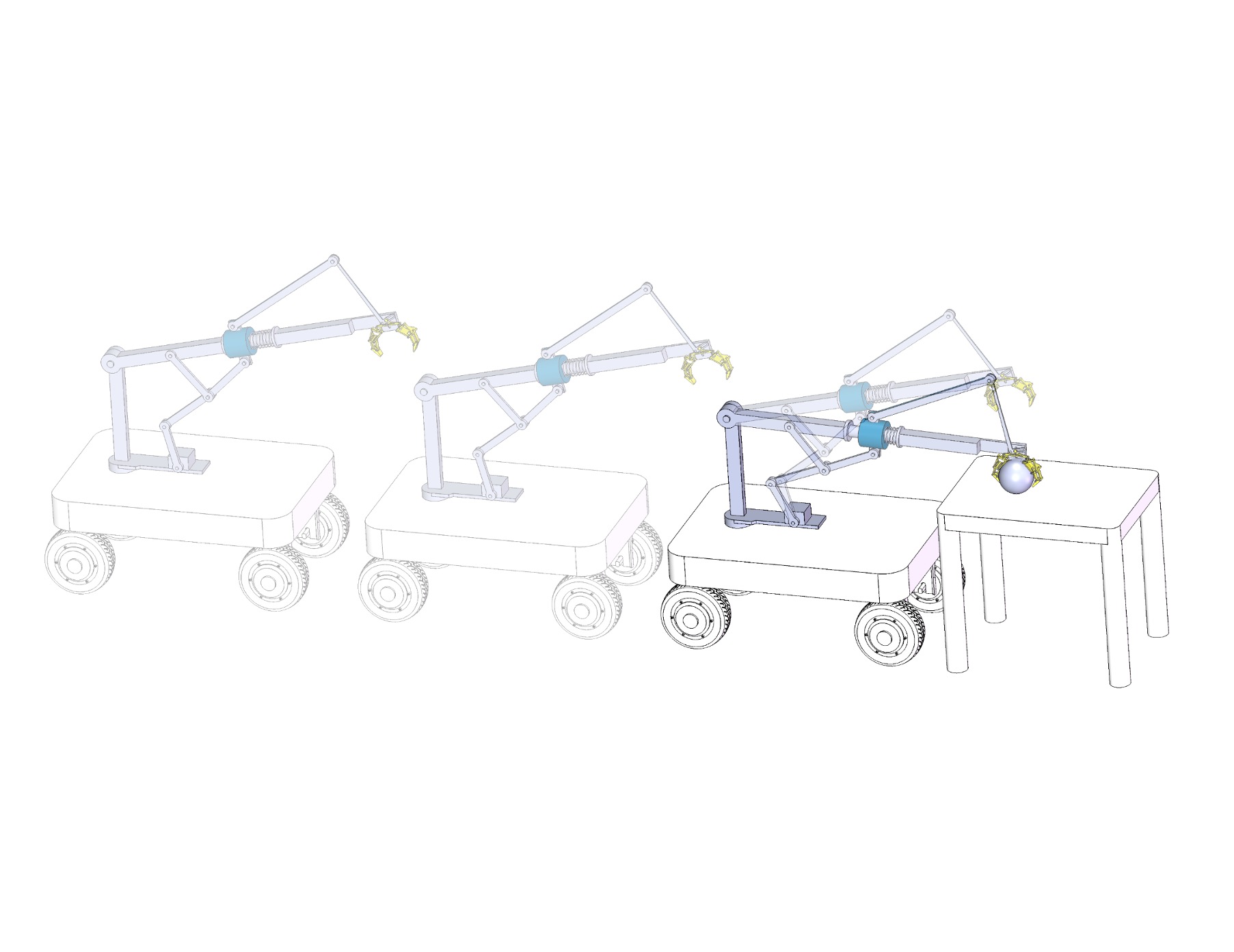}
		\add{\caption{Simulation of coordinated motion between the mobile vehicle and the robotic arm.}
		\label{figure9}}
	\end{figure}

The simulated version of the grasping process is illustrated in Fig. \ref{figure9}. Initially, the trajectory of the end-effector is calculated based on the position of the target object. Eq. \ref{eq22} determines the horizontal displacement of the movable vehicle required to reach the pre-grasping position. Following this, the UMLM executes the grasping sequence as outlined, ensuring synchronized motion and manipulation with the mobile platform. This coordinated execution facilitates precise interaction with the target object, demonstrating the integrated functionality of the UMLM system.

\add{\section{Statics Modeling and Kinetostatics-based Contact Force Analysis of the UMLM Grasping}}

In our efforts to enhance the design and optimization of the UMLM system, we have developed a static modeling of the gripper's fingers. This model serves as a foundational kinetostatics framework for understanding the mechanical interactions that occur during the grasping process. Our kinetostatics-based analysis is centered on the contact forces of each finger that are pivotal to the gripper's performance, as they directly influence the stability and effectiveness of the grasping.
	
To streamline the computational process and focus on the core statics, we have made certain simplifications to neglect the gravitational effects acting on the fingers and the frictional forces that arise between the fingers and the object being grasped, which allows us to isolate and analyze the primary forces without the added complexity of secondary forces.

\add{
Under the assumption of complete grasping with all three joints of each finger in contact with the object, a schematic diagram of the finger’s equivalent cross-sectional mechanism is shown in Fig. \ref{figure10}. The contact points are randomly distributed. The \(\{O_{3}-x_3y_3z_3\}\) coordinate system is established, with the origin at point \(O_{3}\) representing the mid-point of the underactuated metamorphic gripper. The \(x_3\)-axis is aligned with the direction perpendicular to \(l_{9}\), the \(y_3\)-axis is aligned with the direction perpendicular to \(l_{9}\). Other notational conventions are as shown in Fig. \ref{figure10}. 
}
\add{
	\begin{figure}[!htbp]
		\centering 
		\includegraphics[width=11.36cm]{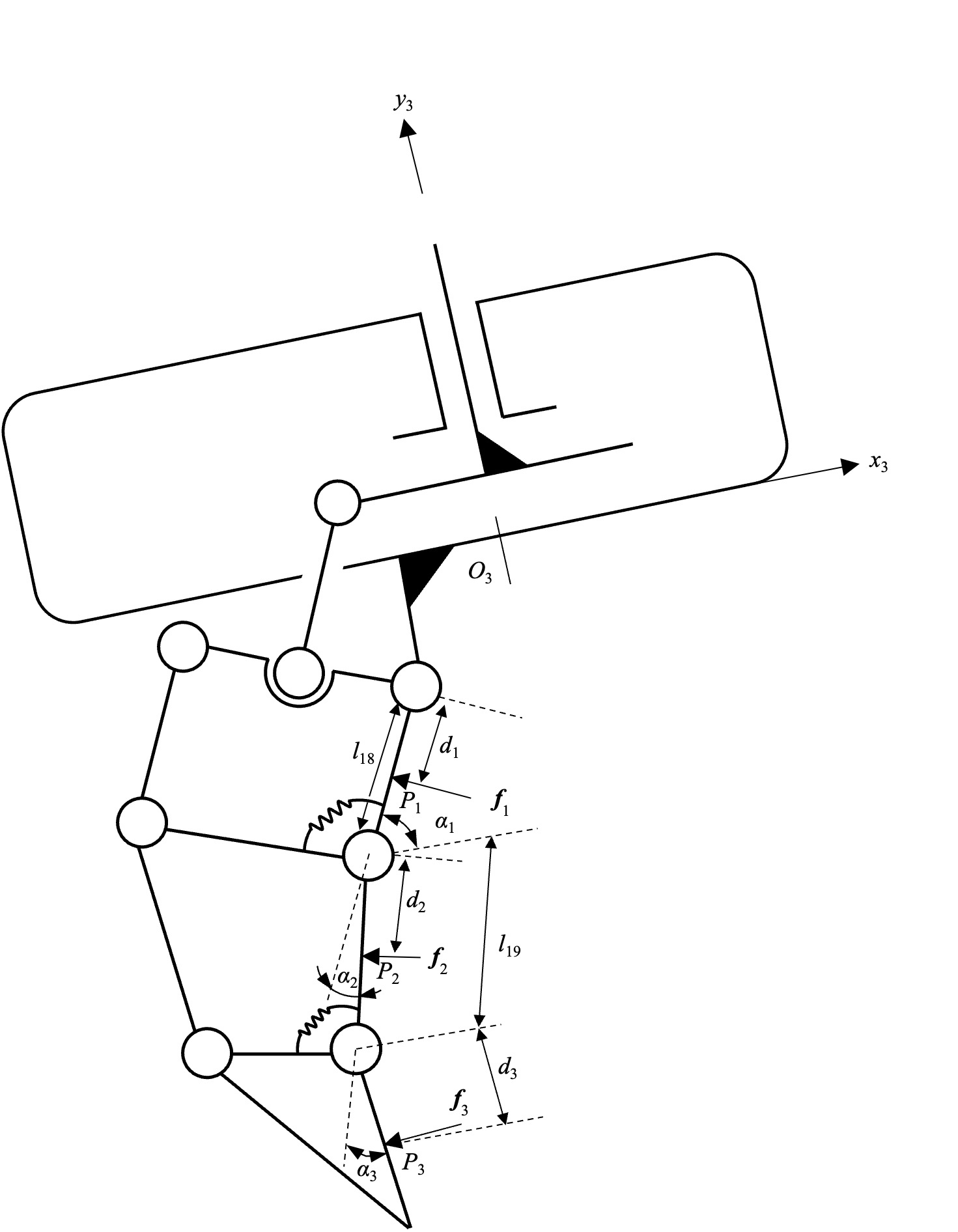}
		\add{\caption{Kinetostatic modeling of the UMLM underactuated finger mechanism.}
		\label{figure10}}
	\end{figure}
}

\add{When the gripper has finished grasping the object, the most common case is that in which each joint of an individual finger has a random contact point with the object. The contact points are designated as \(P_1\), \(P_2\), \(P_3\). \(\boldsymbol{f}_i\) stands for the reaction force of the \(i\)-th joint of a single finger emanating from the object to the segment at point \(P_i\). \(\boldsymbol{\tau}_1\) represents the input torque from the metamorphic arm. \(\boldsymbol{\tau}_2\) and \(\boldsymbol{\tau}_3\) represent the input torque vector by the springs \(s_1\) and \(s_2\). \(\tau_{s_1}\) and \(\tau_{s_2}\) represent the initial torque of the springs \(s_1\) and \(s_2\). \(k_1\) and \(k_2\) represent the stiffness of the torsional springs \(s_1\) and \(s_2\). \(\delta \boldsymbol{\alpha}_1\), \(\delta \boldsymbol{\alpha}_2\), and \(\delta\boldsymbol{\alpha}_3\) represent the difference between the current and initial angles of the \(i\)-th joint. \(\delta \boldsymbol{r}_1\), \(\delta \boldsymbol{r}_2\), \(\delta \boldsymbol{r}_3\) represent the virtual displacements of the three points. \(\alpha_1\), \(\alpha_2\), \(\alpha_3\) denote the angles between the three structures, and \(d_1\), \(d_2\), \(d_3\) are the distances between the contact points and the joints. \(\dot{\alpha}_i\) denote the corresponding angular velocity of \(\alpha_i\).}

The position vector of the three contact points can be expressed as follows:

\add{
\small
\begin{equation}
\begin{aligned}
& \boldsymbol{P}_1=\left(-d_1 \cos \alpha_1,\, -d_1 \sin \alpha_1\right), \\
& \boldsymbol{P}_2=\left(-l_{18} \cos \alpha_1 - d_2 \cos (\alpha_1 + \alpha_2),\, -l_{18} \sin \alpha_1 - d_2 \sin (\alpha_1 + \alpha_2)\right), \\
& \boldsymbol{P}_3=\left(-l_{18} \cos \alpha_1 - l_{19} \cos (\alpha_1 + \alpha_2) - d_3 \cos (\alpha_1 + \alpha_2 + \alpha_3),\right.\\
&\qquad\left. -l_{18} \sin \alpha_1 - l_{19} \sin (\alpha_1 + \alpha_2) - d_3 \sin (\alpha_1 + \alpha_2 + \alpha_3)\right).
\end{aligned}
\end{equation}
}

\add{Differentiating the positions yields the velocities:}
\add{
\begin{equation}
\begin{aligned}
\delta{\boldsymbol r}_1 &= \left( d_1 \dot{\alpha}_1 \sin \alpha_1,\; - d_1 \dot{\alpha}_1 \cos \alpha_1 \right),\\
\delta{\boldsymbol r}_2 &= \left( l_{18} \dot{\alpha}_1 \sin \alpha_1 + d_2 (\dot{\alpha}_1+\dot{\alpha}_2)\sin (\alpha_1+\alpha_2),\;
- l_{18} \dot{\alpha}_1 \cos \alpha_1 - d_2 (\dot{\alpha}_1+\dot{\alpha}_2)\cos (\alpha_1+\alpha_2) \right),\\
\delta{\boldsymbol r}_3 &= \left( l_{18} \dot{\alpha}_1 \sin \alpha_1 + l_{19} (\dot{\alpha}_1+\dot{\alpha}_2)\sin(\alpha_1+\alpha_2)
+ d_3 (\dot{\alpha}_1+\dot{\alpha}_2+\dot{\alpha}_3)\sin(\alpha_1+\alpha_2+\alpha_3),\right.\\
&\qquad \left. - l_{18} \dot{\alpha}_1 \cos \alpha_1 - l_{19} (\dot{\alpha}_1+\dot{\alpha}_2)\cos(\alpha_1+\alpha_2)
- d_3 (\dot{\alpha}_1+\dot{\alpha}_2+\dot{\alpha}_3)\cos(\alpha_1+\alpha_2+\alpha_3) \right).
\end{aligned}
\label{eq7}
\end{equation}
}

The reaction forces emanating from the object, \(\boldsymbol{f}_1, \boldsymbol{f}_2, \boldsymbol{f}_3\) can be expressed as follows:
\add{
	\begin{equation}
		\begin{aligned}
			& \boldsymbol{f}_1=\left(f_1 \sin \alpha_1,-f_1 \cos \alpha_1\right), \\
			& \boldsymbol{f}_2=\left(f_2 \sin(\alpha_1+\alpha_2),-f_2 \cos (\alpha_1+\alpha_2)\right), \\
			& \boldsymbol{f}_3=\left(f_3 \sin(\alpha_1+\alpha_2+\alpha_3),-f_3 \cos(\alpha_1+\alpha_2+\alpha_3)\right).
		\end{aligned}
         \label{eq8}
	\end{equation}
	}
   
\add{The principle of virtual work can be employed to formulate the kinetostatics of compliant mechanisms:}
\add{
	\begin{equation}	\boldsymbol{f}^\mathrm{T}\delta\boldsymbol{r}=\boldsymbol{\tau}^\mathrm{T} \dot{\boldsymbol{\alpha}}.
	\end{equation}
    }

\add{Since $\sin^2 \theta + \cos^2 \theta = 1$, through  Eq.~\ref{eq7} and Eq.~\ref{eq8}, we have
}
\add{
\begin{equation}
\begin{aligned}
\boldsymbol{f}_i^\mathrm{T}\delta\boldsymbol{r}_i
&=
\begin{bmatrix}
{f}_1 & {f}_2 & {f}_3
\end{bmatrix}
\begin{bmatrix}
-d_1 & 0 & 0 \\
-l_{18}\cos\alpha_2 & -d_2 & 0 \\
-l_{18}\cos(\alpha_2+\alpha_3) & -l_{19}\cos\alpha_3 & -d_3
\end{bmatrix}
\begin{bmatrix}
\dot{\alpha}_1 \\
\dot{\alpha_1+\alpha}_2 \\
\dot{\alpha_1+\alpha_2+\alpha}_3
\end{bmatrix}.
\end{aligned}
\end{equation}
}

\add{This leads to the relationship between the forces $f_i$ and $d_i$:}
\add{
	\begin{equation}
		\begin{aligned}
		&f_3 = -\frac{\tau_3}{d_3}, \\
		&f_2 = -\frac{\tau_2}{d_2} +\frac{\tau_3}{d_2} +\frac{\tau_3 l_{19} \cos \alpha_3}{d_3 d_2}, \\
		&f_1 = -\frac{\tau_1}{d_1} + \left( \frac{\tau_2}{d_2} - \frac{\tau_3 l_{19} \cos \alpha_3}{d_3 d_2}-\frac{\tau_3}{d_2} \right) \frac{l_{18} \cos \alpha_2}{d_1} + \frac{\tau_3 l_{18} \cos(\alpha_2+\alpha_3)}{d_3 d_1}+\frac{\tau_3 l_{19} \cos \alpha_3}{d_3 d_1}+\frac{\tau_3}{d_1}.
		\end{aligned}
	\end{equation}	}		

\add{Through further simplification, we get the relationship between $f_1, f_2$ and $f_3$ with $d_1, d_2$ and $d_3$ as follows:}
\add{
\begin{equation}
		\begin{aligned}
        &f_1 = -\frac{1}{d_1} \left(\tau_1 - l_{18} \cos \alpha_2 \left( \frac{\tau_2}{d_2} - \frac{\tau_3 l_{19} \cos \alpha_3}{d_3 d_2}-\frac{\tau_3}{d_2} \right) - \frac{\tau_3 l_{18} \cos (\alpha_2+\alpha_3)}{d_3}-\frac{\tau_3 l_{19} \cos \alpha_3}{d_3}-\tau_3 \right), \\
        &f_2 = -\frac{1}{d_2} \left(\tau_2 -\tau_3- \frac{\tau_3 l_{19} \cos \alpha_3}{d_3} \right),\\
        &f_3 = -\frac{\tau_3}{d_3}.
	\end{aligned}
	\end{equation}
}

\add{The input torque exerted by the springs can be expressed as follows:}
\add{
	\begin{equation}
		\begin{aligned}
			& 
        \tau_2=-(k_1\alpha_2\frac{\pi}{180}+\tau_{{s}_1}), \\
        &
        \tau_3=-(k_2\alpha_3\frac{\pi}{180}+\tau_{{s}_2}).
        		\end{aligned}
	\end{equation}
	}
    
\add{A kinetostatic modeling of the UMLM's grasping can be acquired by these nonlinear equations. If the distances between the contact points and the joints \(d_1\), \(d_2\) and \(d_3\) are specified, the proposed kinetostatic modeling can enable computation of the reaction forces of the manipulator gripper through numerical iterations.}
	
\begin{figure}[htbp]
		\centering 
		\includegraphics[width=11.7cm]{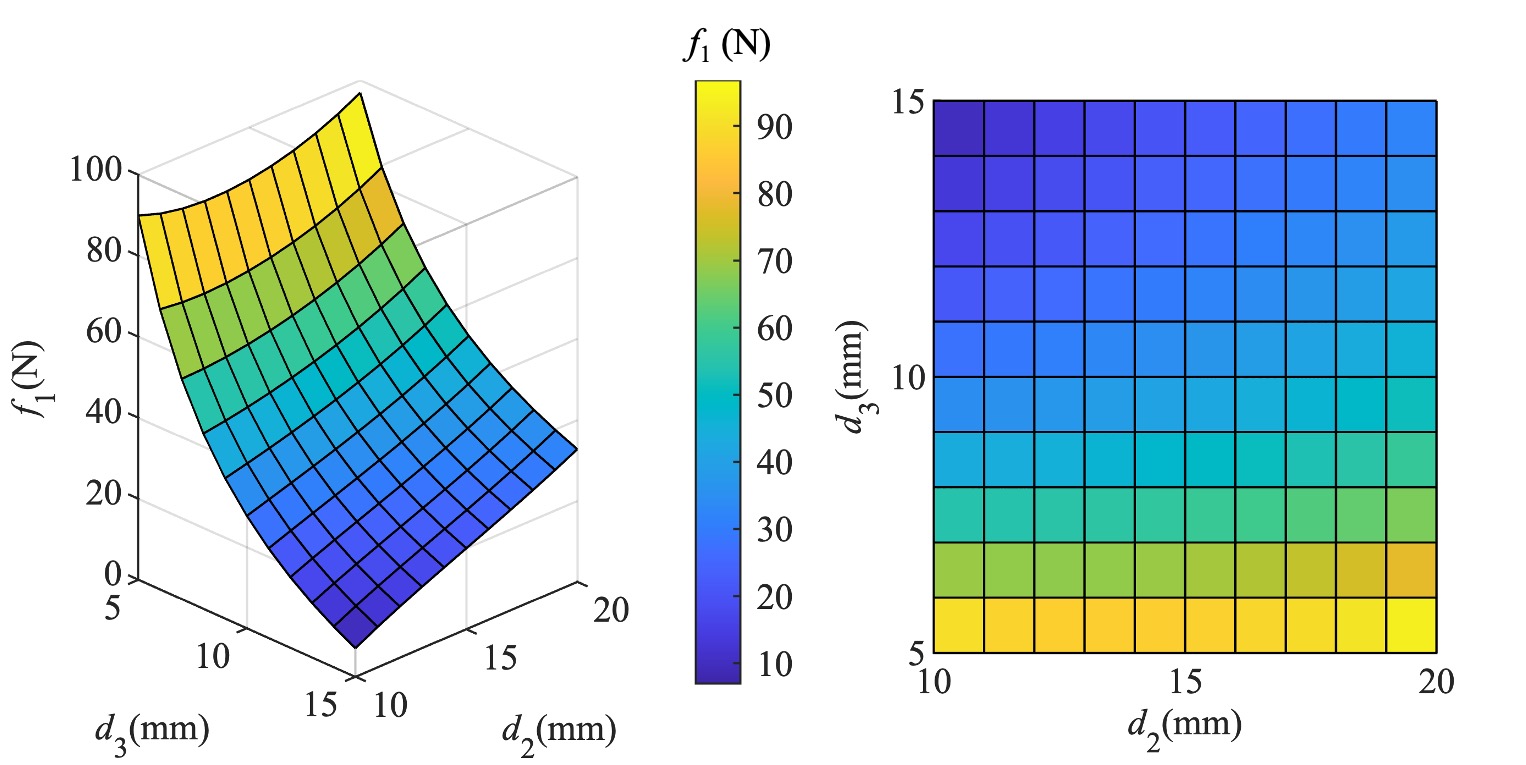}
		\add{\caption{The relationship between \(f_1\), \(d_2\), and \(d_3\).}
		\label{figure11}}
	\end{figure}

\begin{figure}[htbp]
		\centering 
\includegraphics[width=11.7cm]{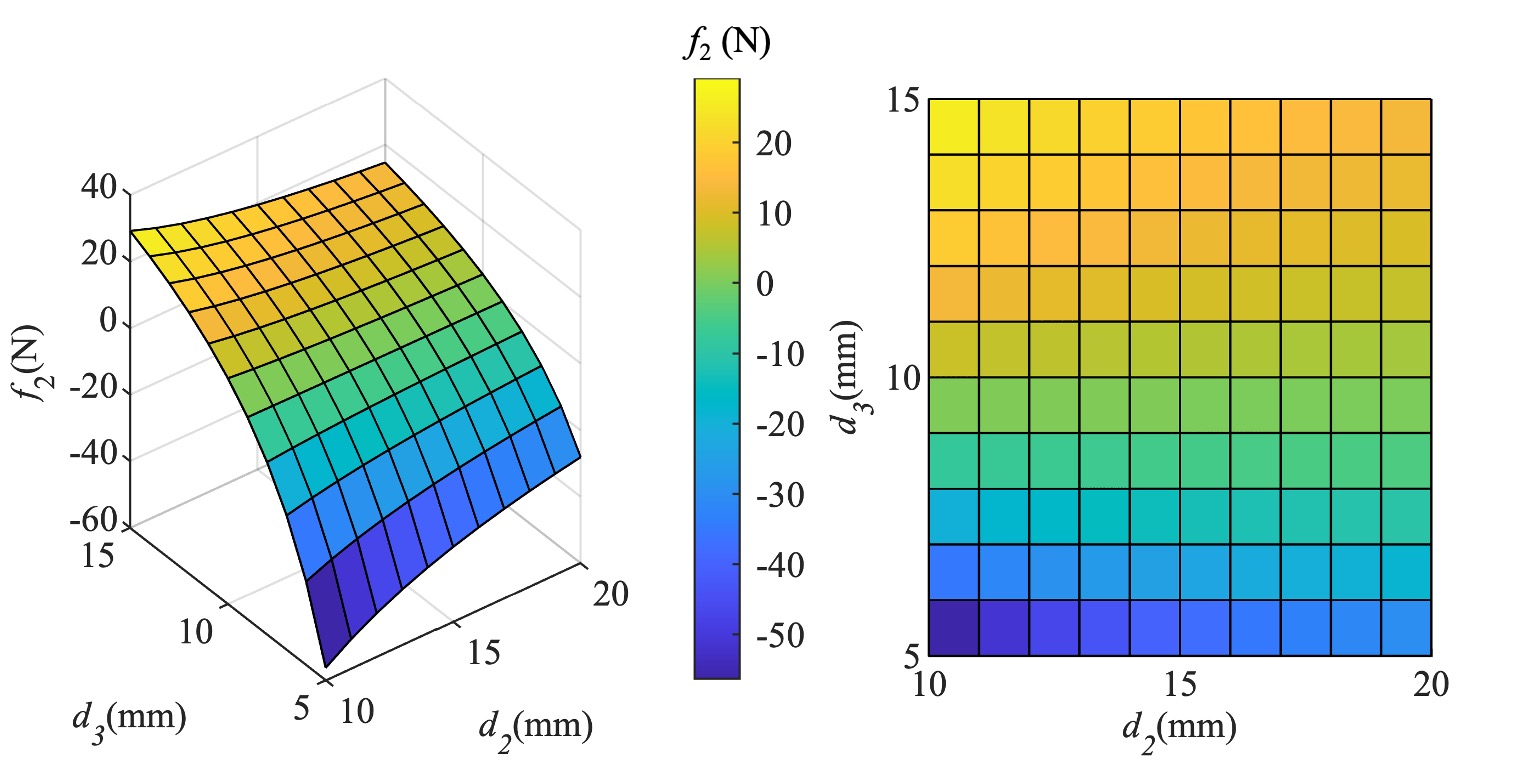}
		\add{\caption{The relationship between \(f_2\), \(d_2\), and \(d_3\).}
		\label{figure12}}
	\end{figure}

\add{Figs. ~\ref{figure11} and ~\ref{figure12} provide a visual representation of the dynamic interplay between the forces \(f_1\), \(f_2\) and the geometry of the gripper system. The projection of the three-dimensional diagram of forces can be seen in the two-dimensional diagram on the right side of Figs.~\ref{figure11} and ~\ref{figure12}. As illustrated in Fig.~\ref{figure11}, \(f_1\) is positively related to \(d_2\) and negatively related to \(d_3\). As illustrated in Fig.~\ref{figure12}, \(f_2\) is positively related to \(d_2\) and \(d_3\).} This trend is indicative of the mechanical behavior of the gripper segments in response to the changing contact point positions. To further elaborate, when the contact point is closer to the end of the gripper, the pressure exerted by the second and third segments on the object increases significantly, which is crucial for securing the object within the gripper. Conversely, the pressure from the first segment diminishes, shifting its role from being a primary force applicator to maintaining the structural integrity of the gripper. This shift aligns with the gripping process, where the first segment's role is to support the gripper's shape, allowing the second and third segments to apply the necessary pressure for a secure grip.

    
In conclusion, the results presented in Figs.~\ref{figure11} and~\ref{figure12} are invaluable for further refinement of the gripper's design and for informing the development of similar robotic systems that require precise and adaptable gripping capabilities. This observed behavior underscores the rationality and efficacy of the gripper's design. The ability of the gripper to adapt its force distribution based on the contact point positions ensures that the gripper can accommodate a variety of object shapes and sizes while maintaining a secure grip.
	
\section{Particle-Swarm Optimization of Gripper Dimensional Parameters}
\subsection{Formulation of Objective Function}
The primary objective of the gripper mechanism is to ensure a reliable and robust grasping function, which is essential for various applications in robotics and automation. This functionality is predicated on the ability to generate sufficient gripping force under the constraints of limited driving power. The challenge lies in striking a delicate balance: the fingers must apply adequate contact forces to securely grasp the object without exerting excessive pressure that could lead to damage. This requirement underscores the need for precision in the design and control of the gripper mechanism.
	
In an ideal stable grasping scenario, the gripper mechanism must maintain a uniform distribution of contact forces at the finger joints. Uniformity in contact forces is crucial for several reasons. Firstly, it ensures that the object is held securely without any risk of slippage or misalignment. Secondly, it helps to prevent localized stress concentrations that could lead to damage to either the gripper or the object being grasped. Lastly, a uniform force distribution contributes to the longevity and reliability of the gripper mechanism by reducing wear and tear on its components.
	
In addition to the uniform distribution of forces, the design of the finger mechanism should prioritize compactness and favorable force transmission characteristics. A compact design is not only aesthetically pleasing but also practical, as it allows for the gripper to be integrated into spaces with limited room for maneuver. 
	
Favorable force transmission characteristics are equally important, as they ensure that the forces generated by the driving mechanism are effectively transmitted to the fingers without significant loss or distortion. \add{This efficiency maximizes the gripper's performance with the available driving power and ensures that the gripper can operate effectively across a wide range of grasping scenarios.}
	
In line with these design requirements, the gripper mechanism should be engineered to achieve a uniform distribution of contact forces across all segments.
	
Let:

	\begin{equation}
		\begin{aligned}
		&f_{\min} = \min(f_1, f_2, f_3), \\
		&f_{\max} = \max(f_1, f_2, f_3),
		\end{aligned}
	\end{equation}
	
\noindent{then, the objective function of mechanism parameter optimization is:}

	\begin{equation}
		\Phi(x) = |f_{\min} - f_{\max}|.
	\end{equation}

Up to this point, we have obtained the objective function and its parameter settings. In the next section, we will carry out optimization using the PSO algorithm based on this objective function.
	
\subsection{Gripper Dimension Optimization via Particle-Swarm Optimization Algorithm}
The optimization process is important in the design and refinement of complex systems, and in this study, we have leveraged the Particle-Swarm Optimization (PSO) algorithm from the MATLAB Optimization Toolbox to achieve our optimization goals. The PSO algorithm is a powerful computational method that simulates the social behavior of a swarm of particles, where each particle represents a potential solution to the problem at hand. These particles collaboratively explore the search space, adjusting their positions and velocities in response to both their own experiences and the collective intelligence of the swarm.
	
The PSO algorithm is renowned for its efficiency in the resolution of complex, nonlinear, and multidimensional optimization challenges. It operates by iteratively refining the solutions, thereby gradually converging towards the global optimum. This is achieved through the interplay of two main components: the personal best solution of each particle, which is the best solution it has encountered, and the global best solution, which is the best solution found by any particle in the swarm.

\add{When the transmission pressure angle of the mechanism is zero, the force transmission efficiency is the highest. To ensure optimal force transmission across the entire gripping range of the finger mechanism, we conduct the optimization under this state with the relative angles between the knuckles being \(45^\circ\). We assume the drive torque \(\tau_1 = 1000 \, \text{Nmm}\), \(l_{18} = 38.3 \, \text{mm}\), \(l_{19} = 30 \, \text{mm}\), \(l_{24} = 20.5 \, \text{mm}\). The objective function \(\Phi\) could be obtained through calculation, where \(x = (l_{16}, l_{21}, l_{22}, k_1, k_2, \tau_{s_1}, \tau_{s_2})\) represents the optimization parameters. For our specific application, we have defined the search space boundaries for the optimization parameters. The lower bounds (\(lb\)) and upper bounds (\(ub\)) for the parameters were set as follows:\\
\(lb=[20, 10, 10, 10, 10, 0, 0]\) and \(ub = [30, 15, 15, 1000, 1000, 200, 200]\).} These bounds define the limits within which the particles can search for the optimal solution, ensuring that the solutions remain feasible and practical.

To ensure a comprehensive search and to enhance the likelihood of finding a near-optimal solution, we have set the number of particles in the swarm to 1000. This large swarm size allows for a more exhaustive exploration of the search space, increasing the diversity of solutions and reducing the risk of converging prematurely to a suboptimal solution.
	
\begin{figure}[htbp]
	\centering 
	\includegraphics[width=0.65\linewidth]{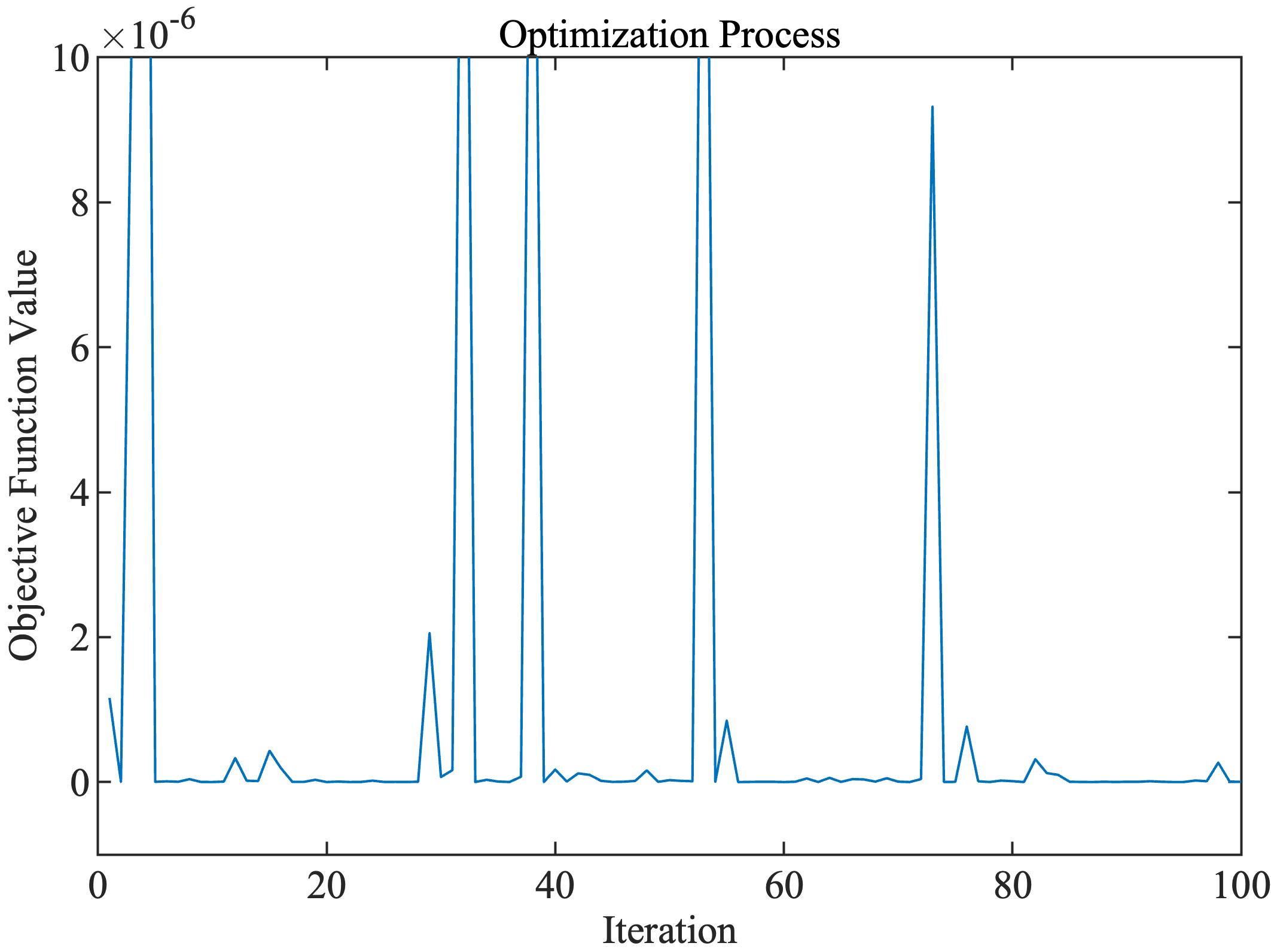}
	\caption{Particle-Swarm Optimization process.}
	\label{figure13}
	\end{figure}

Through multiple runs of the PSO algorithm, we have generated a robust set of solutions. A total of 100 distinct solution sets have been identified and are presented in Fig.~\ref{figure13}. These solutions represent the outcomes of our optimization efforts, offering a range of potential configurations that meet the specified criteria and constraints.
		
The results from the PSO algorithm not only provide us with a set of optimized parameters but also contribute to a deeper understanding of the underlying relationships and dynamics within the system. By analyzing the solutions presented in Table ~\ref{tab:gripper_params}, we can gain insights into the performance of the gripper mechanism under various conditions and make informed decisions to further refine its design.
	
Comparing the objective function values of the group with the worst performance in parameter optimization and that of the group with the best performance, the result of the objective function shows that $\Phi_a$ is $0.0247$ and $\Phi_b$ is $3.8508 \times 10^{-11}$, where:		
	\[x_a = (26.78, 15.00, 13.35, 12.1, 525.6, 186.56, 199.43),\]
	\[x_b = (28.02, 15.00, 13.58, 346.5, 794.1, 184.43, 196.29).\]	
This complies with the conditions for the effectiveness of the uniform stress Particle-Swarm Optimization process and fulfills the design requirements.
	
%% Use a table environment to create tables.
%% Refer following link for more details.
%% https://en.wikibooks.org/wiki/LaTeX/Tables
\begin{table}[t]%% placement specifier
%% Use tabular environment to tag the tabular data.
%% https://en.wikibooks.org/wiki/LaTeX/Tables#The_tabular_environment
\centering%% For centre alignment of tabular.
%% Use \caption command for table caption and label.
\caption{Worst and best group parameters of the gripper.}\label{tab:gripper_params}
\begin{tabular}{l c c}%% Table column specifiers
\hline
Parameter & Group A & Group B \\
\hline
\add{$l_{16}$ (mm)} & 26.78 & 28.02 \\
\add{$l_{21}$ (mm)} & 15.00 & 15.00 \\
\add{$l_{22}$ (mm)} & 13.35 & 13.58 \\
\add{$k_1$ (Nmm/rad)} & 12.1 & 346.5 \\
\add{$k_2$ (Nmm/rad)} & 525.6 & 794.1 \\
\add{$\tau_{s_1}$ (Nmm)} & 186.56 & 184.43 \\
\add{$\tau_{s_2}$ (Nmm)} & 199.43 & 196.29 \\
Objective Function $\Phi$ & 0.0247 & $3.8508 \times 10^{-11}$ \\
\hline
\end{tabular}
\end{table}
	
\section{Simulation of Adaptive Grasping for the UMLM and Discussion of Results}
\begin{figure}[htbp]
		\centering 		\includegraphics[width=13cm]{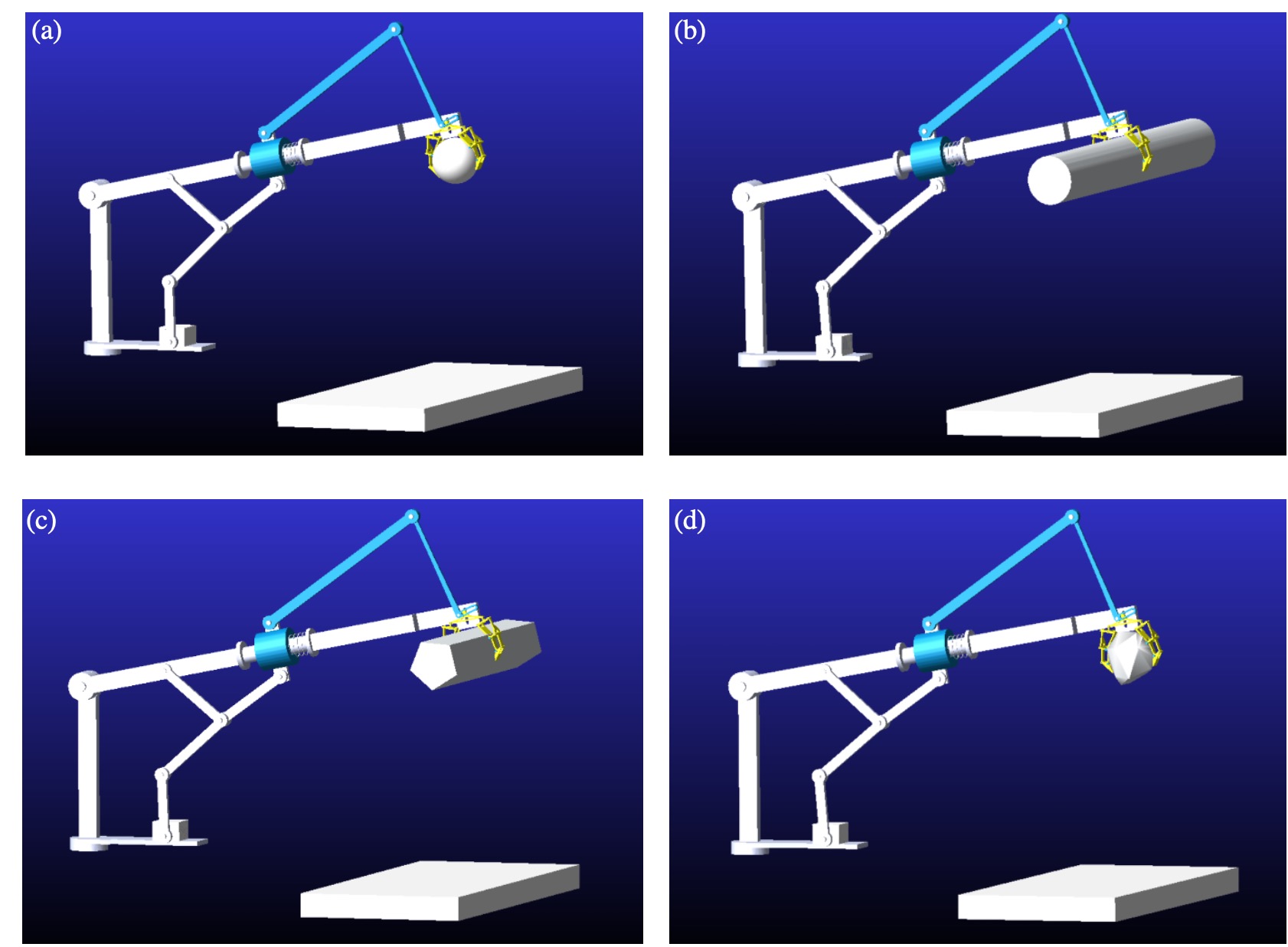}
 \add{\caption{Simulation validations of UMLM's grasping capability of different objects. (a) Sphere grasping simulation. (b) Cylinder grasping simulation. (c) Pentagonal prism grasping simulation. (d) Irregular object grasping simulation.}
		\label{figure15}}
	\end{figure}

\add{To substantiate the theoretical modeling delineated in this paper, the 3D simulation model of the UMLM was constructed in Adams, adhering to the principle of metamorphosis engineered in this research. A series of simulations were executed by grasping different objects involving sphere, cylinder, pentagonal prisms with single-sided asymmetric structures and irregular objects with randomly generated geometries as demonstrated in Fig.~\ref{figure15}.} The primary objective of the grasping trials was to monitor its performance in securing target objects that fall within a specific volume and mass spectrum. The results of these simulations provide insights into the UMLM's capabilities and limitations for potential improvement, paving the way for the successful implementation of the UMLM in various applications.
	
\add{The simulation framework incorporated key parameters to ensure physically accurate system behavior. Material properties were defined based on the 6061 aluminum alloy with a stiffness of \(3.33 \times 10^7\) N/m to simulate component deformation under operational loads. Contact dynamics utilized an extended Lagrangian formulation with a force exponent of 1.50. Coulomb friction was employed to model tangential forces, with static and dynamic friction coefficients of 0.2 and 0.1. This parameter configuration ensured energy-conservative interactions during object manipulation while maintaining numerical stability. Kinematic pairs were established to constrain motion relationships, while collision conditions were configured using the solid-to-solid contact model detailed in Table~\ref{tab:simulation_params} to prevent unrealistic penetrations.}

\add{The actuation system employed a 120W DC motor, providing the necessary torque for manipulation tasks. To accurately evaluate the manipulation capabilities of the underactuated hand, a series of dynamic simulations were conducted in ADAMS using representative objects frequently encountered in real-world scenarios, including spheres, cylinders, and prisms. The spherical objects had a diameter of 100 mm and a mass of 0.524 kg, whereas the cylindrical objects had a base diameter of 100 mm, a length of 600 mm, and a mass of 4.94 kg. Prisms are representative polyhedral shapes frequently encountered in industrial applications. In the simulation, the pentagonal prism had a side length of 64mm with a mass of 0.9kg.}

\begin{figure}[htbp]
  \centering  \includegraphics[width=1\linewidth]{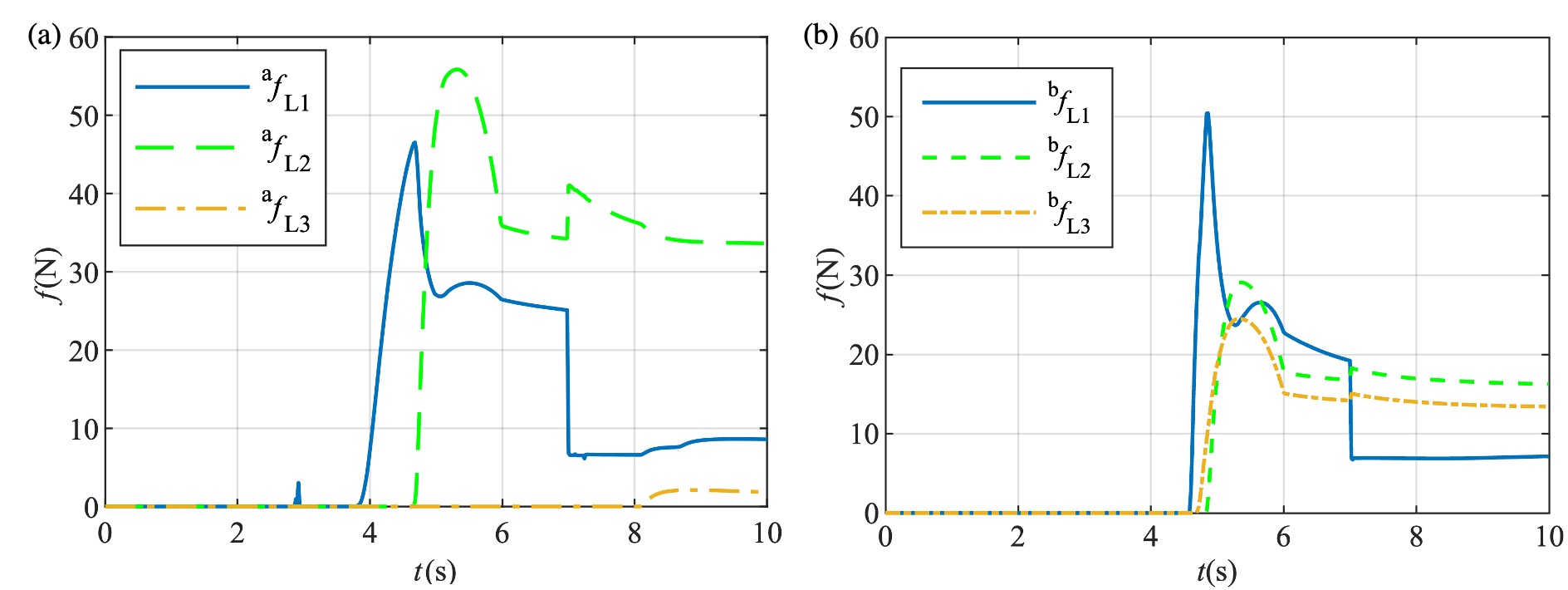}
  \add{\caption{Contact forces of unilateral knuckles when grasping the sphere. (a) Contact forces of unilateral knuckles of the worst group. (b) Contact forces of unilateral knuckles of the best group.}
  \label{figure16}}
\end{figure}

When the UMLM is commanded to grab spherical objects such as footballs, the contact forces of unilateral knuckles are shown in Fig.~\ref{figure16}. We can observe that after the grasping action is completed in the 7th second, the contact forces \( ^af_{L1} \), \( ^af_{L2} \) and \( ^af_{L3} \) in parameter group (a) are uneven, which is demonstrated in the simulation video attached in supplement video. As shown in the experimental results of Fig.~\ref{figure16}, the optimized parameter group (b), after being processed by the PSO algorithm, exhibits a more uniform distribution of the three-finger grasping contact forces \( ^bf_{L1} \), \( ^bf_{L2} \) and \( ^bf_{L3} \) during the grasping process. In contrast, the grasping force of the worst parameter group (a) is more dispersed. This result validates the effectiveness of the PSO algorithm optimization.

Furthermore, we can observe that when grasping spherical objects of the same size and weight, the optimized parameter group (b) exerts a decreased three-finger grasping force compared to group (a). This demonstrates that optimized parameters can effectively enhance grasping efficiency and reduce energy consumption.

\begin{figure}[htbp]
  \centering   \includegraphics[width=1\linewidth]{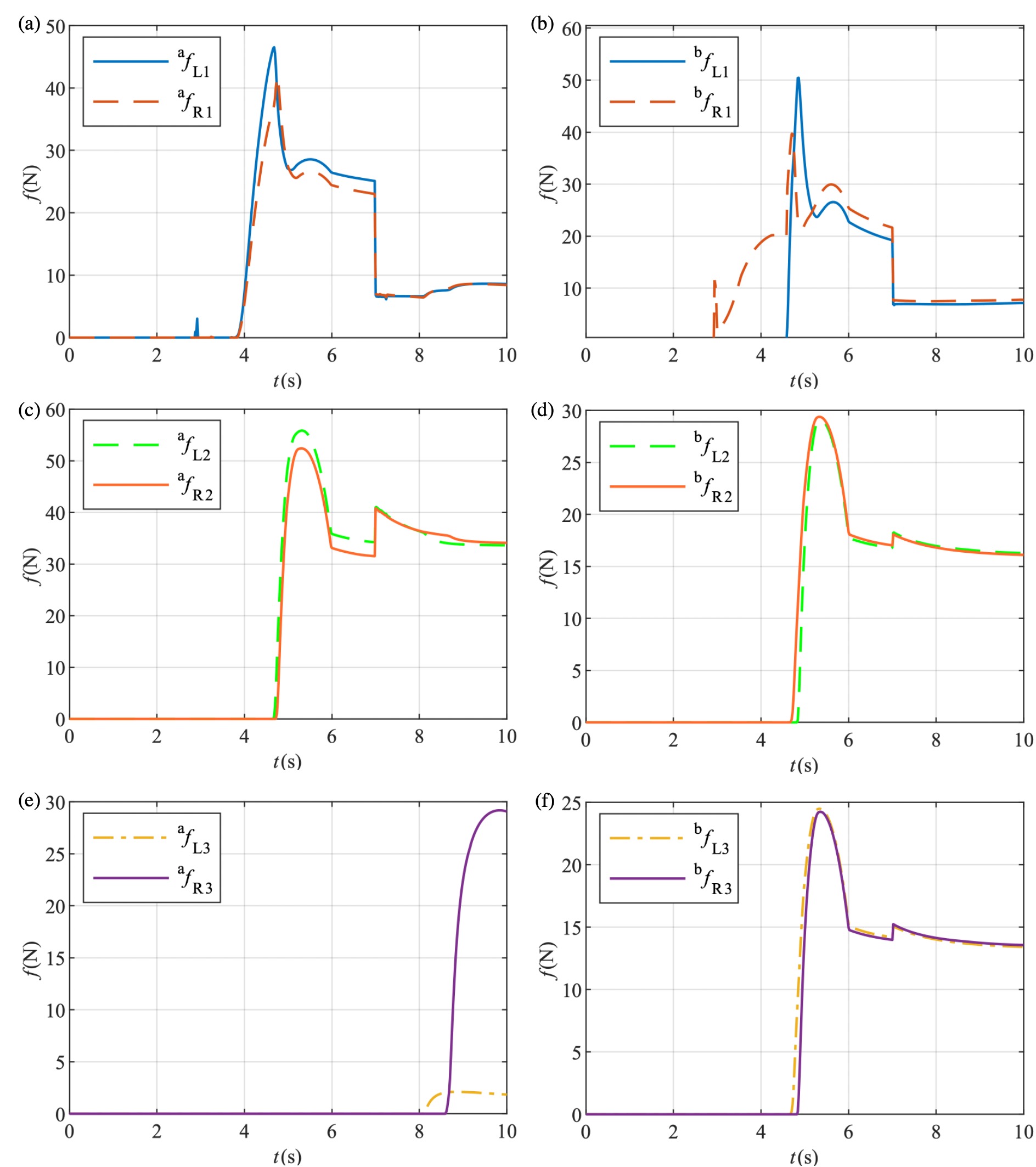}
  \add{\caption{Contact forces of bilateral knuckles during sphere grasping process. (a)-(b) Contact forces 1 of bilateral knuckles of the worst and the best group. (c)-(d) Contact forces 2 of bilateral knuckles of the worst and the best group. (e)-(f) Contact forces 3 of bilateral knuckles of the worst and the best group.}
  \label{figure17}}
\end{figure}

Additionally, the experiment compared the contact forces of the bilateral knuckles produced during motion. From Fig.~\ref{figure17}, we can see that the contact forces of the symmetrical bilateral knuckles with the sphere also tend to be uniform in group (b), which verifies the feasibility and optimization of our metamorphic mechanism design and the applicability of the PSO algorithm we used.

\begin{figure}[htbp]
  \centering 
  \includegraphics[width=1\linewidth]{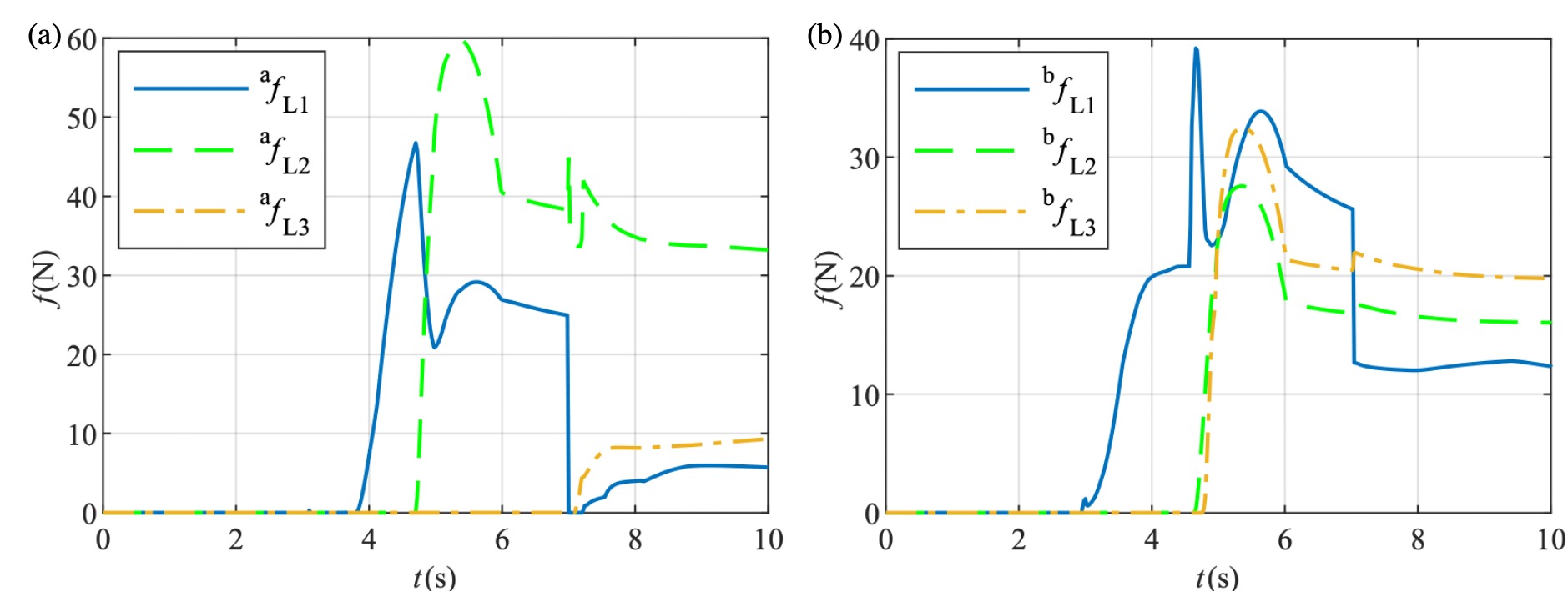}
  \add{\caption{Contact forces of unilateral knuckles when grasping the cylinder. (a) Contact forces of unilateral knuckles of the worst group. (b) Contact forces of unilateral knuckles of the best group.}
  \label{figure18}}
\end{figure}

The motion trajectory of enveloping and grasping cylindrical objects such as pieces of rebar is depicted in Fig.~\ref{figure18}.  An important observation regarding the distribution of forces among the knuckles of the UMLM can be made from Fig.~\ref{figure18}. It is evident that the force exerted on three knuckles of group (b) is more uniform compared to that on group (a) and the grasping efficiency of group (b) is higher than that of group (a).

\begin{figure}[htbp]
  \centering 
  \includegraphics[width=1\linewidth]{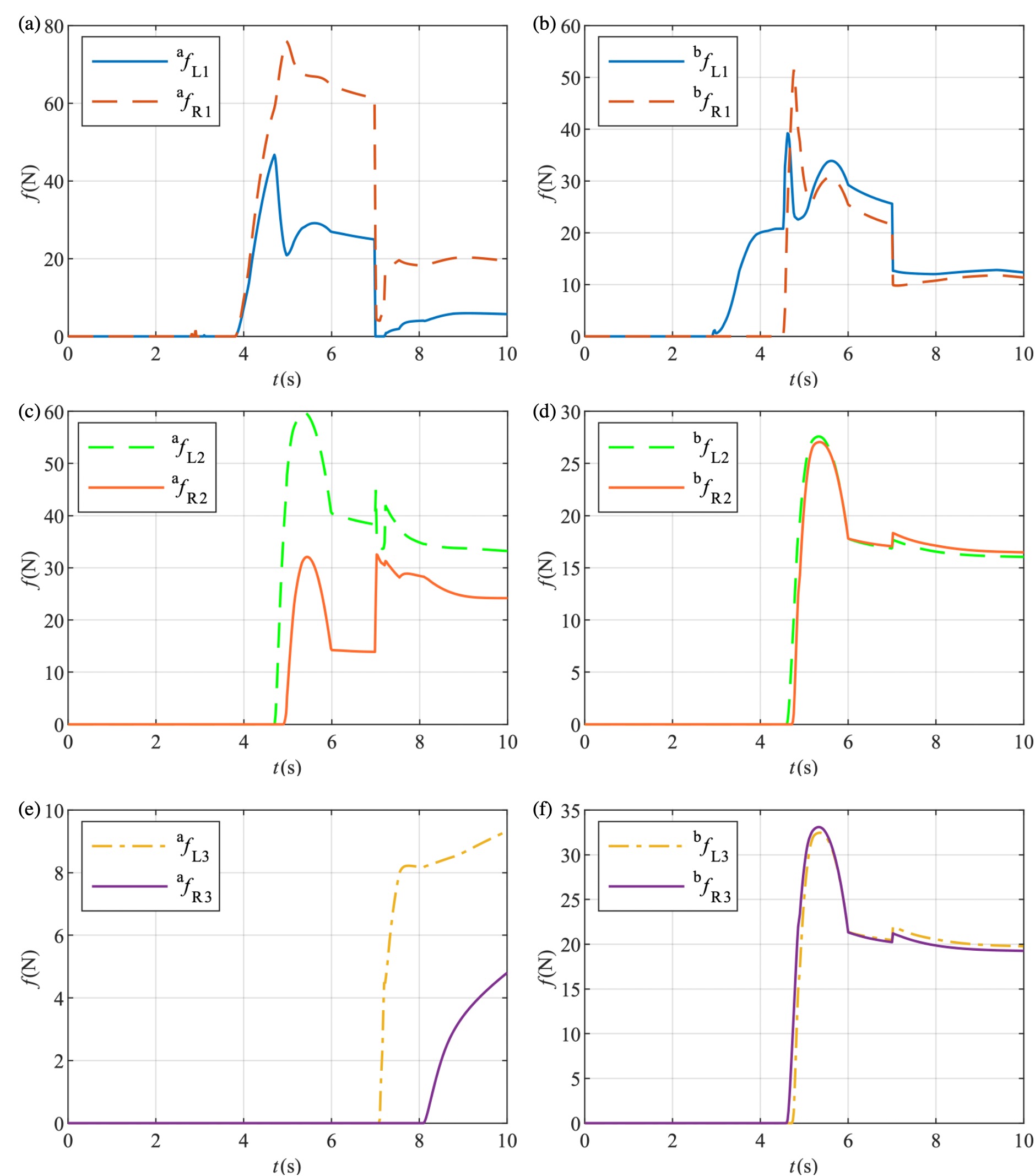}
  \add{\caption{Contact forces of bilateral knuckles during cylinder grasping process. (a)-(b) Contact forces 1 of bilateral knuckles of the worst and the best group. (c)-(d) Contact forces 2 of bilateral knuckles of the worst and the best group. (e)-(f) Contact forces 3 of bilateral knuckles of the worst and the best group.}
  \label{figure19}}
\end{figure}

Furthermore, the experimental analysis was extended to compare the different contact forces exerted by the bilateral knuckles throughout the motion sequence. As depicted in Fig.~\ref{figure19}, it is observed that the contact forces at the symmetrical bilateral knuckles with the cylindrical object exhibit a tendency towards uniformity. This observation serves as empirical validation for the efficacy and optimization of the metamorphic mechanism design, as well as the suitability of the Particle-Swarm Optimization (PSO) algorithm employed in this study.

The simulation experiments conducted with the underactuated hand have yielded promising results, effectively validating the feasibility and stability of the proposed metamorphic mechanism. These experiments have demonstrated the UMLM's capability to adapt to a variety of grasping scenarios, which is a testament to the design's versatility and robustness. By comparing the simulation data results of group (a) and group (b) to each other, we can observe that group (b), which is optimized by the PSO algorithm, can achieve stable grasping of objects with a smaller force. This validates the effectiveness of PSO in terms of reducing power consumption. 

These figures indicate that the UMLM are not only capable of adapting to different objects but also of applying the necessary force to securely grasp and manipulate them. The simulation results highlight the UMLM's strength in terms of the maximum grasping force it can exert. It suggests that the UMLM are well-suited for tasks that require a combination of precision and strength, such as in assembly lines, logistics, or any environment where objects need to be moved or manipulated with adaptability and efficiency.
	
\section{Conclusions}
This paper introduces a pioneering innovative UMLM, showcasing its exceptional grasping force and adaptability across a variety of object geometries. The kinetostatics-based contact force analyses provide insights into the UMLM's real-time structural morphing and interaction dynamics, laying a robust theoretical foundation for the design and simulation. To further optimize the performance of the gripper, Particle-Swarm Optimization (PSO) has been effectively utilized to refine the size parameters of the gripper's structure. This optimization ensures robust adaptability across various applications, highlighting the UMLM's potential in scenarios that demand efficient and adaptable loading solutions. The simulation corroborates the simplicity, versatility, self-adaptability, and strong grasping ability of the UMLM, with the potential to significantly reduce costs and enhance performance in robotics applications. Beyond the specific prototype, the proposed modeling and optimization framework offers generalizability to a wider class of underactuated metamorphic manipulators. This work lays a foundation for scalable, cost-effective robotic systems capable of adaptive loading, with promising implications for future deployment as a versatile tool in industrial automation, service robotics, and complex manipulation tasks in uncertain settings.

\section*{Acknowledgments}
We acknowledge the support of the Key Program of the National Natural Science Foundation of China (Grant 52335003), the Guangdong S\&T program (Grant 2023ZT10Z002), and the Shenzhen Science and Technology Program (Grant KQTD20240729102052065), the A*STAR under its RIE2025 Manufacturing, Trade and Connectivity(MTC) Industry Alignment Fund-Pre-Positioning (IAF-PP) funding scheme (Project No. SC29/24-814711-MRMR). We are grateful to Su Jiarui for proofreading the kinematic descriptions and formulas of the metamorphic mechanism, as well as for carefully checking and improving the manuscript.

\section*{Competing interests}
All authors certify that they have no affiliations with or involvement in any organization or entity with any financial interest or nonfinancial interest in the subject matter or materials discussed in this manuscript.

\end{document}